\documentclass[journal]{IEEEtran}

\usepackage{amsmath,amsfonts}
\usepackage{algorithmic}
\usepackage{algorithm}
\usepackage{array}
\usepackage[caption=false,font=normalsize,labelfont=sf,textfont=sf]{subfig}
\usepackage{textcomp}
\usepackage{stfloats}
\usepackage{url}
\usepackage{verbatim}
\usepackage{graphicx}
\usepackage{cite}
\usepackage{pifont}
\usepackage[hidelinks]{hyperref}
\begin{document}

\title{Demographics-Informed Neural Network for Multi-Modal Spatiotemporal forecasting of Urban Growth and Travel Patterns Using Satellite Imagery}
\author{%
  Eugene Kofi Okrah Denteh, 
  Andrews Danyo, 
  Joshua Kofi Asamoah, 
  Blessing Agyei Kyem, 
  Armstrong Aboah
  \IEEEmembership{Member, IEEE}\\[1ex]
  \thanks{Manuscript created 29 May, 2025; This work was developed by Eugene Kofi Okrah Denteh, Andrews Danyo, Joshua Kofi Asamoah, Blessing Agyei Kyem and Armstrong Aboah. The authors are with the Department of Civil, Construction and Environmental Engineering, North Dakota State University, Fargo, ND 58102 USA. (email: eugene.denteh@ndsu.edu; andrews.danyo@ndsu.edu; joshua.asamoah@ndsu.edu; blessing.agyeikyem@ndsu.edu; armstrong.aboah@ndsu.edu)}%
}

\markboth{Preprint}{Denteh \MakeLowercase{\textit{et al.}}: Demographics-Informed Neural Network …}




\maketitle

\begin{abstract}
This study presents a novel demographics informed deep learning framework designed to forecast urban spatial transformations by jointly modeling geographic satellite imagery, socio-demographics, and travel behavior dynamics. The proposed model employs an encoder-decoder architecture with temporal gated residual connections, integrating satellite imagery and demographic data to accurately forecast future spatial transformations. The study also introduces a demographics prediction component which ensures that predicted satellite imagery are consistent with demographic features, significantly enhancing physiological realism and socioeconomic accuracy. The framework is enhanced by a proposed multi-objective loss function complemented by a semantic loss function that balances visual realism with temporal coherence. The experimental results from this study demonstrate the superior performance of the proposed model compared to state-of-the-art models, achieving higher structural similarity (SSIM: 0.8342) and significantly improved demographic consistency (Demo-loss: 0.14 versus 0.95 and 0.96 for baseline models). Additionally, the study validates co-evolutionary theories of urban development, demonstrating quantifiable bidirectional influences between built environment characteristics and population patterns. The study also contributes a comprehensive multimodal dataset pairing satellite imagery sequences (2012-2023) with corresponding demographic and travel behavior attributes, addressing existing gaps in urban and transportation planning resources by explicitly connecting physical landscape evolution with socio-demographic patterns. 
\end{abstract}


\section{Introduction}\label{Intro}
Effective transportation infrastructure forms the critical foundation upon which modern urban development thrives; connecting communities to essential services and catalyzing economic growth \cite{gramlich94,POKHAREL2023100817,SERDAR2022103452}. As cities expand, well-designed transportation networks not only facilitate the efficient movement of people and goods \cite{DANYO2025} but also reshape urban landscapes, creating opportunities for new commercial and residential zones while alleviating congestion and enhancing quality of life \cite{su15108410, WAN2024384}. Recognizing these interconnected dynamics, current transportation planning approaches have established a direct relationship between socio-demographics, travel behavior and spatial representations, enabling planners to better understand how population characteristics influence travel patterns and infrastructure needs across diverse landscapes \cite{WANG201725,wang2024deep}. As a result, transportation planning has become an increasingly complex, data-driven process as it involves integrating demographic factors and travel behavior features such as trip purpose and mode choice to accurately forecast demand and make informed decisions\cite{denteh2025integrating}. This approach has enabled planners optimize resource allocation and develop efficient mobility networks customized for diverse community needs. 

Despite adopting a data-centric approach to transportation planning, effectively modeling the dynamic temporal relationship that exists between the built environment, socio-demographics and travel behavior remains a persistent challenge \cite{LIU2024}. Several studies have highlighted this constraint, indicating that addressing this problem could significantly enhance the effectiveness of transportation planning. For instance, Acheampong and Silva \cite{racheampong}, found that conventional static models inadequately capture the evolving interdependencies between demographic shifts, spatial changes, and travel behavior due to their inability to account for temporal variability. Similarly, Kim et al. \cite{kim2024travel} also described the significant shortcomings in regional travel demand forecasting models, emphasizing their limited capability to adapt dynamically to spatial changes revealed through satellite imagery or evolving socio-demographic contexts. This limitation stems primarily from the use of time-invariant methodologies in current transportation models, which fail to capture the dynamic temporal relationships between the built environment, socio-demographics, and travel behavior. These time-invariant approaches manifest in several ways across transportation planning frameworks. For instance, conventional static traffic assignment models operate under the assumption that travel demand remains uniformly distributed throughout analysis periods, obscuring peak demand patterns. Similarly, the widely used four-step transportation planning methodology employs fixed origin-destination matrices that cannot adapt to evolving travel patterns \cite{boyles2006comparison}. Central to this problem is the lack of a diverse dataset, sequentially curated to reflect the progressive changes of spatial representations with time \cite{vanetten2021multitemporalurbandevelopmentspacenet}. This has often led to infrastructure wastage where existing infrastructure is demolished, in some cases, to accommodate new projects addressing evolving transportation needs. A prime example occurred in Rochester, New York, where a section of the Inner Loop freeway, which was constructed in the late 1960s based on static traffic projections, was demolished in 2017 after actual vehicle volumes remained far below forecasts \cite{pugh2024rochester}. This costly transition to a street-level boulevard highlights how reliance on fixed modeling techniques can lead to underused infrastructure and expensive retrofits once urban mobility needs evolve.

To address these limitations, this study proposes a novel Demographics-Informed deep Neural Network (DINN) designed to model geographic satellite imagery, demographic and travel behavior dynamics simultaneously. To achieve the above, the current study employs an encoder-decoder architecture, incorporating temporal gated residual connections, capable of jointly processing sequential satellite imagery and corresponding demographic data for counties across the United States. This enables the model to predict future spatial representations accurately, capturing subtle and gradual geographic transformations. 
To this end, the study makes the following contributions;
\begin{enumerate}

    \item The introduction of a spatial prediction framework that jointly processes nationwide satellite image sequences and demographic data of counties across the United States through temporal gated residual connections to predict future spatial representations depicting the changes that can occur within a geographic area.
    \item The development of a demographics-informed approach for predicting urbanization patterns, ensuring that predicted satellite images maintain both physiological realism and statistical alignment with socio-demographic data. 
    \item The development of a demographic prediction network that effectively extracts sociodemographic information from satellite imagery, demonstrating the ability to infer population characteristics, age distributions, racial composition, and educational attainment from physical landscape features.
    \item The introduction of a travel behavior prediction network that capitalizes on encoded feature representations learned from the demographic prediction network, establishing the strong correlation that exists between demographics and travel behavior. 
    \item The study also proposes a multi-objective loss function that ensures visually realistic and temporally coherent predictions, significantly improving the perceptual and quantitative accuracy of generated images. We also introduce a semantic loss function that aligns the semantic representation output from the demographic and travel behavior prediction network decoders with the encoded bottleneck features, preserving critical latent information during the decoding process. 
    \item The use of temporal difference heatmaps to visually track urbanization patterns across multi-year sequences. The framework combines structural similarity metrics (SSIM) with demographic validation through synthetic image regression, enabling joint assessment of spatial accuracy and socioeconomic plausibility in predicted urban developments.
    \item The development of a novel multimodal dataset that pairs sequential satellite imagery with corresponding demographic and travel behavior attributes. This dataset comprises temporally ordered satellite images spanning 2012 to 2023 across all 50 United States, aligned with comprehensive demographic indicators and travel behavior metrics including age groups, sex, race and educational attainment.
\end{enumerate}

\section{Literature Review}
\subsection{Evolution of Transportation Planning Approaches}

Transportation planning has undergone a fundamental transformation from isolated infrastructure development to integrated multimodal strategies that acknowledge the complex interdependencies between transportation systems and urban development. This evolution began in the pre-1960s era, which prioritized accommodating rapid car ownership growth through large-capacity highway construction, often neglecting broader community impacts \cite{Sciara03072017,HORCHER2021100196}. Building upon this foundation, the 1960s-1980s period marked a significant methodological shift with the introduction of the four-step model (trip generation, distribution, modal split, traffic assignment), which provided systematic traffic prediction frameworks but paradoxically reinforced the separation of transportation from land use planning \cite{Mladenovic,BIBRI2020100021}. This methodological isolation proved problematic, as contemporary approaches since the 1990s have increasingly recognized the need for integrated strategies that emphasize sustainability, social equity, and demand management through mechanisms such as congestion pricing and multimodal integration \cite{BALSAS20151}. Most recently, the field has witnessed the integration of artificial intelligence and machine learning technologies for intelligent transportation systems. These advanced computational approaches represent a fundamental departure from traditional planning methodologies, with applications spanning from real-time traffic optimization and predictive maintenance to autonomous vehicle integration and comprehensive urban mobility management \cite{abirami2024systematic}.

\subsection{Challenges in Dynamic Urban Transportation Planning}

The limitations identified in historical transportation planning approaches have intensified in contemporary dynamic urban environments, where the pace of change far exceeds the adaptive capacity of traditional planning models. Dynamic transportation planning faces critical challenges from rapidly evolving landscapes and unpredictable urban growth patterns \cite{racheampong}, creating a feedback loop where planning inadequacies exacerbate the very problems they seek to address. Compounding these temporal challenges, planning models struggle with real-time changes in population density and travel behavior influenced by technological advances and economic shifts \cite{kim2024travel}. Several studies have introduced innovative image processing approaches in transportation engineering research \cite{kyem2024advancing,kyem2024pavecap,kyem2024weather,kyem2025context,asamoah2025saam,dontoh2025visual}.
Current research further highlights the complexity of spatiotemporal urban dynamics, demonstrating that conventional approaches systematically fail to capture the multiscale nature of urban development processes \cite{li2019,zhang2024characteristics}. This temporal-spatial disconnect manifests in infrastructure solutions that are fundamentally misaligned with ground realities, creating a costly cycle of retrofitting and premature demolition \cite{boyles2006comparison}. Historical precedents underscore the severity of these consequences, exemplified by Interstate 95's construction through Miami's Overtown neighborhood, which displaced 10,000-12,000 residents and destroyed 40 blocks of community infrastructure \cite{karas2015highway}.
These interconnected challenges underscore an urgent need for machine learning models that can transcend the limitations of current approaches by simultaneously capturing holistic urban development patterns and integrating spatiotemporal complexity with demographic dynamics in real-time adaptive frameworks.

\section{Approach}
This section begins with the problem definition to establish the mathematical foundations, followed by a detailed description of the proposed network architecture that processes both the spatial representations and demographic inputs in parallel, and conclude with the training methodology that unifies these components into a cohesive predictive system capable of forecasting spatial transformations with quantifiable confidence metrics. Following this, we present our travel behavior prediction network, which extends the model's capabilities to capture transportation dynamics.

\subsection{Problem Formulation}
Despite the significant advancements in data-driven transportation planning methodologies, effectively modeling the dynamic and temporal relationship between geographic changes in satellite imagery and socio-demographic patterns remains a persistent challenge. This limitation has hindered progress in addressing the unpredictable nature of rapid urbanization and evolving commuting preferences, resulting in infrastructure wastage; which has to do with existing infrastructure being demolished, in some cases, to accommodate new
projects to address the evolving transportation needs. To address this challenge, this paper proposes the utilization of satellite image sequences and corresponding demographics to predict future spatial representations and also predict the future demographics and travel behavior.

To effectively model the relationship between geographic changes in satellite images and demographic factors, let \(\{x_{t-n},\dots,x_{t}\}\) denote a temporal sequence of \((n+1)\) historical satellite images, where each image \(x_{i}\in\mathbb{R}^{H\times W\times C}\) comprises RGB channels, \(C\) with spatial dimensions \(H\times W\) (height and width, respectively). Correspondingly, let \(\{d_{t-n},\dots,d_{t}\}\) represent the associated demographic feature vectors, each \(d_{i}\in\mathbb{R}^{f}\) containing \(f\) socio-demographic variables. The objective is to predict the future satellite image \(\hat{x}_{t+1}\in\mathbb{R}^{3\times H\times W}\) and the future demographic vector \(\hat{d}_{t+1}\in\mathbb{R}^{f}\) by using the temporal sequence of past satellite images, \(\{x_{t-n},\dots,x_{t}\}\) together with their corresponding demographics, \(\{d_{t-n},\dots,d_{t}\}\) as input into a network that predicts the future year. The general methodology adopted to achieve the objective is described in section \ref{AO}.

Extensive research has established that a strong relationship exists between demographic characteristics and travel behavior patterns \cite{wang2024deep,LU19991,MWALE2022100683}. To quantify and leverage this established relationship in our predictive framework, we formalize the demographic-travel behavior correlation as:

\begin{equation}
\mathcal{C}(d_i, t_i) = \mathbb{E}[\langle \phi(d_i), \psi(t_i) \rangle_{\mathcal{H}}] > \tau
\end{equation}

where $\phi: \mathbb{R}^{f} \to \mathcal{H}$ and $\psi: \mathbb{R}^{m} \to \mathcal{H}$ are feature mapping functions into a reproducing kernel Hilbert space $\mathcal{H}$, $\langle \cdot, \cdot \rangle_{\mathcal{H}}$ denotes the inner product in $\mathcal{H}$, and $\tau$ represents a correlation threshold. This correlation enables the establishment of a bidirectional predictive relationship:

\begin{equation}
t_i = \mathcal{F}_{d \to t}(d_i) + \epsilon_t, \quad d_i = \mathcal{F}_{t \to d}(t_i) + \epsilon_d
\end{equation}

where $\mathcal{F}_{d \to t}: \mathbb{R}^{f} \to \mathbb{R}^{m}$ and $\mathcal{F}_{t \to d}: \mathbb{R}^{m} \to \mathbb{R}^{f}$ are nonlinear mapping functions, and $\epsilon_t, \epsilon_d$ represent modeling residuals.

The primary objective is to jointly predict the future satellite image $\hat{x}_{t+1}\in\mathbb{R}^{H\times W\times C}$, the future demographic vector $\hat{d}_{t+1}\in\mathbb{R}^{f}$, and the future travel behavior vector $\hat{t}_{t+1}\in\mathbb{R}^{m}$ by leveraging the temporal sequences of past observations. This tri-modal prediction problem is formulated as the optimization:
{\small
\begin{equation}
\begin{aligned}
  (\hat{x}_{t+1}, \hat{d}_{t+1}, &\hat{t}_{t+1}) = \arg\min_{x,d,t}\; \mathcal{L}_{spat}(x,x_{t+1})
  +\\ \alpha\,\mathcal{L}_{demo}(d,d_{t+1}) 
  &+ \beta\,\mathcal{L}_{travel}(t,t_{t+1})
  + \gamma\,\mathcal{L}_{cons}(d,t)
\end{aligned}
\end{equation}
}

subject to the constraint:
\begin{equation}
\|\mathcal{C}(\hat{d}_{t+1}, \hat{t}_{t+1}) - \mathcal{C}(d_{t+1}, t_{t+1})\| \leq \delta
\end{equation}

where $\mathcal{L}_{spat}$, $\mathcal{L}_{demo}$, and $\mathcal{L}_{travel}$ represent loss functions for spatial, demographic, and travel behavior predictions respectively, $\mathcal{L}_{cons}$ enforces demographic-travel behavior correlation consistency, $\alpha, \beta, \gamma$ are weighting hyperparameters, and $\delta$ is a tolerance parameter for correlation preservation.

To establish the demographic-travel behavior correlation empirically, we introduce a novel cross-modal embedding framework. Let $\mathcal{E}_d: \mathbb{R}^{f} \to \mathbb{R}^{k}$ and $\mathcal{E}_t: \mathbb{R}^{m} \to \mathbb{R}^{k}$ be demographic and travel behavior encoders respectively, mapping to a shared $k$-dimensional latent space. The correlation is then quantified through the canonical correlation coefficient:

\begin{equation}
\rho(\mathcal{E}_d(d_i), \mathcal{E}_t(t_i)) = \frac{\text{Cov}(\mathcal{E}_d(d_i), \mathcal{E}_t(t_i))}{\sqrt{\text{Var}(\mathcal{E}_d(d_i))\text{Var}(\mathcal{E}_t(t_i))}}
\end{equation}

The general methodology adopted to achieve this multi-objective optimization integrates a demographics-informed neural architecture that processes spatial-temporal satellite imagery while maintaining consistency between demographic projections and travel behavior predictions through shared latent representations and cross-modal regularization terms, as described in section \ref{AO}.

\subsection{Demographics-Informed Satellite Image Prediction Architecture Overview}\label{AO}
\begin{figure*}[!h]
    \centering
    \includegraphics[scale=0.95]{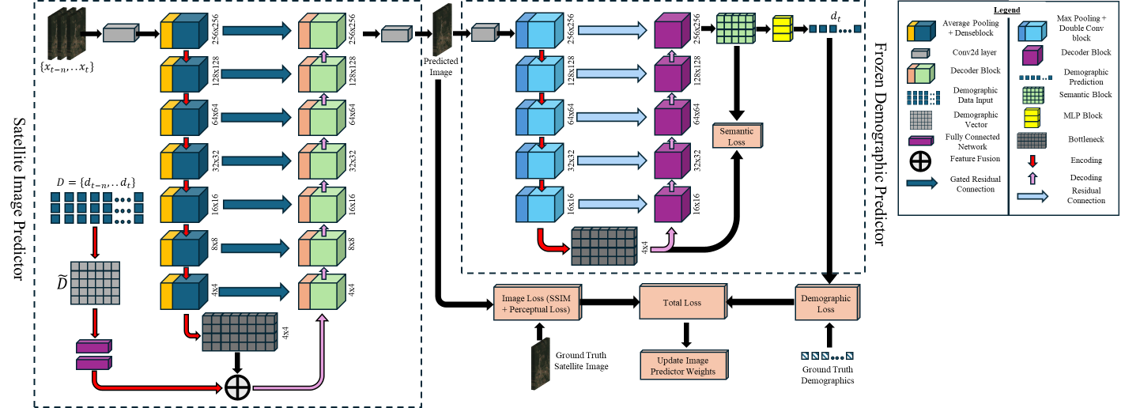}
    \caption{Overall Architecture of Demographics Informed Neural Network (DINN)}
    \label{fig:DINN}
\end{figure*}

The proposed demographics-informed satellite image prediction architecture consists of two interconnected neural networks. This design enables bidirectional information flow between satellite imagery and demographic data, ensuring that predictions maintain both spatial coherence and demographic consistency. The key innovation lies in how these networks communicate: the first network (Satellite Image Predictor) uses a temporal sequence of past satellite images together with the corresponding demographic data sequence to forecast future satellite images, while the second network (Demographic Predictor) validates that these generated images accurately reflect their underlying demographic features. This is illustrated in figure \ref{fig:DINN}.

\subsubsection{Satellite Image Predictor}

\begin{figure}
    \centering
    \includegraphics[scale=0.46]{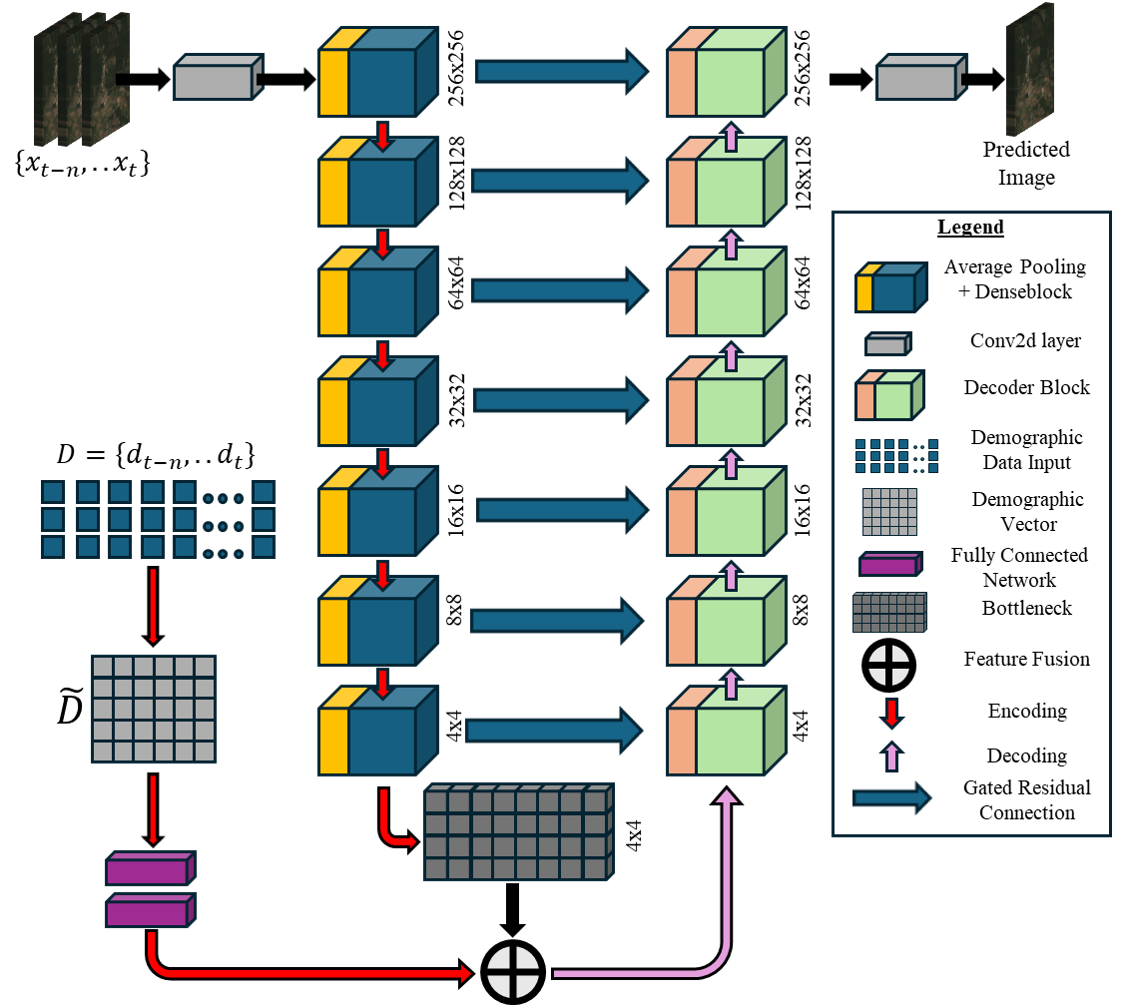}
    \caption{Satellite Image Prediction Architecture with Gated Residual Connections}
    \label{fig:3}
\end{figure}

The Satellite Image Predictor module, as illustrated in figure \ref{fig:3}, implements an encoder-decoder architecture augmented with gated residual connections. This module employs a dual encoding approach consisting of a sequential image encoder for processing the sequence of input satellite images and a demographic encoder for processing corresponding demographic input sequences.

\paragraph{\textbf{Sequential Image Encoder}}

The encoder processes concatenated temporal imagery through a sequence of DenseBlocks that implement dense connectivity patterns for enhanced feature reuse. The input satellite image sequence $\{x_{t-n}, \ldots, x_t\}$ is first reshaped into a single tensor:

\begin{equation}
I = \text{reshape}(E) \in \mathbb{R}^{H\times W\times 3(n+1)}
\end{equation}

where $I$ represents the concatenated input tensor combining all temporal frames along the channel dimension.

Each DenseBlock $\ell$ implements the core dense connectivity principle:

\begin{equation}
x_\ell = H_\ell([x_0, x_1, \ldots, x_{\ell-1}])
\end{equation}

where $x_\ell$ denotes the output feature maps at layer $\ell$, $H_\ell$ is the composite transformation function, and $[x_0, x_1, \ldots, x_{\ell-1}]$ represents the concatenation of feature maps from all preceding layers. The composite transformation $H_\ell$ consists of batch normalization, ReLU activation, and convolution operations applied sequentially.

The growth rate parameter $k$ controls feature expansion, with each DenseBlock contributing $k$ additional feature maps. This creates a hierarchical representation where feature channels expand from 64 to 512 through seven consecutive DenseBlocks, while spatial dimensions progressively reduce through average pooling operations between blocks.

The resulting encoder generates hierarchical feature maps $\{F_1, F_2, \ldots, F_7\}$ where each $F_i \in \mathbb{R}^{H/2^i \times W/2^i \times c_i}$ with $c_i = 64 \cdot 2^{i-1}$, capturing both fine-grained textures and global spatial context.

\paragraph{\textbf{Demographic Encoder}}

The demographic encoder processes the temporal sequence of demographic features through feature-wise normalization followed by nonlinear projection. For the demographic sequence $D = \{d_{t-n}, \ldots, d_t\}$ where each $d_i \in \mathbb{R}^f$, normalization is applied to address scale heterogeneity:

\begin{equation}
\hat{d}_{i,j} = \frac{d_{i,j} - \mu_j}{\sigma_j + \epsilon}
\end{equation}

where $d_{i,j}$ represents the $j$-th demographic feature at time $i$, $\mu_j$ and $\sigma_j$ are the mean and standard deviation of feature $j$ across the temporal sequence, and $\epsilon = 10^{-8}$ provides numerical stability.

The normalized features are temporally concatenated to form a unified vector:

\begin{equation}
\widetilde{D} = [\hat{d}_{t-n}, \ldots, \hat{d}_t] \in \mathbb{R}^{(n+1)f}
\end{equation}

A multi-layer perceptron with dropout regularization projects this high-dimensional vector into a latent embedding $e_d \in \mathbb{R}^{512}$, creating a compact representation that encodes temporal socioeconomic patterns and their interdependencies.

\paragraph{\textbf{Feature Fusion and Gated residual Connections}}

The feature fusion mechanism integrates spatial and demographic information at the bottleneck level. The bottleneck features $F_{bottleneck} \in \mathbb{R}^{h \times w \times 512}$ represent the most abstract spatial representation, where $h$ and $w$ are the reduced spatial dimensions at the bottleneck.

To incorporate demographic context, the demographic embedding is spatially broadcasted:

\begin{equation}
E_d = \text{broadcast}(e_d) \in \mathbb{R}^{h \times w \times 512}
\end{equation}

The fusion operation combines these representations through channel-wise concatenation followed by dimensionality reduction:

\begin{equation}
F_{fused} = \text{Conv}_{1\times1}([F_{bottleneck}, E_d]) \in \mathbb{R}^{h \times w \times 512}
\end{equation}

where $[F_{bottleneck}, E_d]$ denotes channel-wise concatenation and $\text{Conv}_{1\times1}$ represents a 1×1 convolution for dimensionality reduction.

The gated residual connections enable adaptive information flow during decoding. At each decoder level $i$, the gating mechanism evaluates feature relevance:

\begin{equation}
G_i = \sigma(f_{\theta}([E_i, \text{Upsample}(D_{i+1})]))
\end{equation}

where $E_i$ represents encoder features at level $i$, $D_{i+1}$ represents upsampled decoder features from the previous level, $f_{\theta}$ is a 1×1 convolutional layer with batch normalization, $\sigma$ is the sigmoid activation function, and $G_i \in [0,1]$ represents the computed gate weights.

The gated residual connection selectively transmits encoder information:

\begin{equation}
F_{residual,i} = [G_i \odot E_i, \text{Upsample}(D_{i+1})]
\end{equation}

where $\odot$ denotes element-wise multiplication, enabling the network to adaptively emphasize relevant spatial features while suppressing noise based on contextual decoder information.

\paragraph{\textbf{Decoder Pathway}}

The decoder pathway progressively reconstructs spatial information through seven decoding blocks using transposed convolutions. Each decoder level $i$ processes the gated residual connection features:

\begin{equation}
D_i = h_i(F_{residual,i})
\end{equation}

where $h_i$ represents the convolutional operations including batch normalization and ReLU activation at decoder level $i$. This structure systematically restores spatial detail while preserving semantic content from deeper layers.

The final output layer transforms the decoded features into the predicted satellite image:

\begin{equation}
\hat{x}_{t+1} = \tanh(\text{Conv}_{1 \times 1}(D_1)) \in \mathbb{R}^{H\times W\times 3}
\end{equation}

where the $\tanh$ activation ensures pixel values are bounded in the range $[-1, 1]$.

\subsubsection{Demographic Predictor}

\begin{figure}[!h]
    \centering
    \includegraphics[scale=0.47]{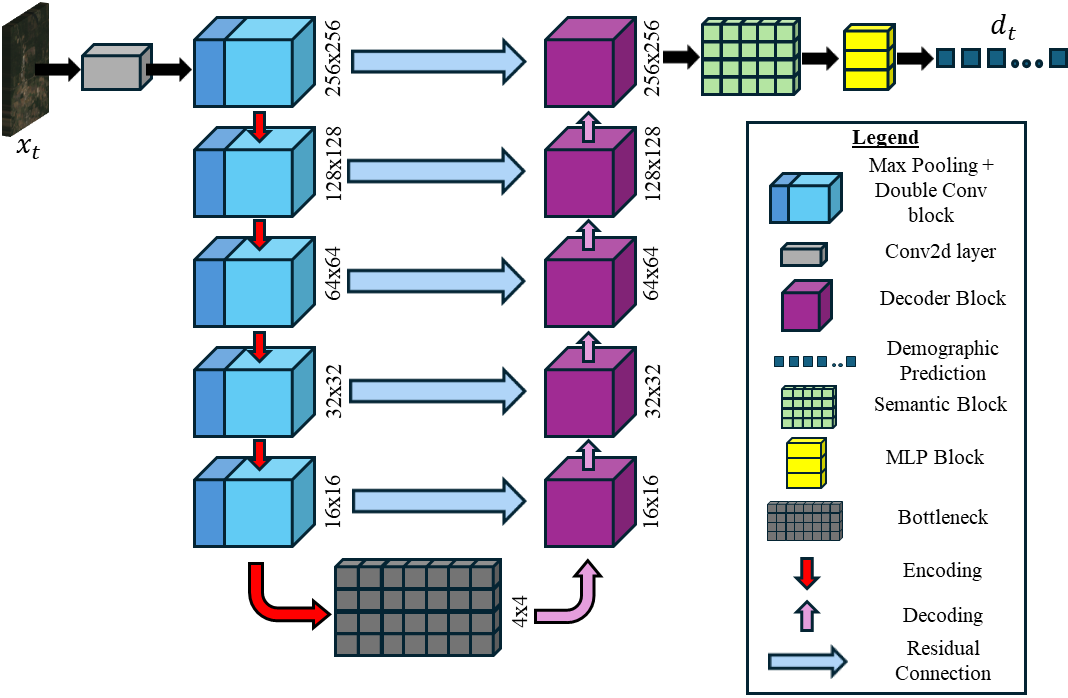}
    \caption{Demographic Prediction Network Architecture}
    \label{fig:2}
\end{figure}

The demographic prediction network also implements an encoder-decoder architecture for demographic regression as illustrated in figure \ref{fig:2}. The encoder pathway consists of four encoding blocks:

\begin{equation}
\begin{split}
    E^{(d)}_i = \text{MaxPool}_{2 \times 2}(\text{Conv}_{3 \times 3}\\(\text{ReLU}(\text{Conv}_{3 \times 3}(E^{(d)}_{i-1}))))
\end{split}
\end{equation}

with feature dimensions progressing as $\{64, 128, 256, 512\}$. The bottleneck representation is:

\begin{equation}
B^{(d)} = \text{Conv}_{3 \times 3}(\text{ReLU}(\text{Conv}_{3 \times 3}(E^{(d)}_4)))
\end{equation}

The decoder pathway implements symmetric upsampling:

\begin{equation}
\begin{split}
    D^{(d)}_i = \text{Conv}_{3 \times 3}(\text{ReLU}(\text{Conv}_{3 \times 3}\\([\text{Upsample}(D^{(d)}_{i+1}), E^{(d)}_{5-i}])))
\end{split}
\end{equation}

The semantic representation pathway extracts global context:

\begin{equation}
S^{(d)} = \text{AdaptiveAvgPool}(B^{(d)}) \in \mathbb{R}^{512}
\end{equation}

The final demographic prediction combines spatial and semantic information:

\begin{equation}
\hat{d}_{t+1} = W_{demo} \cdot [\text{Flatten}(D^{(d)}_1), S^{(d)}] + b_{demo}
\end{equation}

where $W_{demo} \in \mathbb{R}^{f \times (H \times W \times 64 + 512)}$.

\subsection{Travel Behavior Prediction Network}
The Travel Behavior Predictor leverages the frozen demographic encoder to establish demographic-travel behavior correlations. Let $f_e^{frozen}: \mathbb{R}^{H \times W \times 3} \to \mathbb{R}^{h' \times w' \times c'}$ denote the frozen encoder. The travel behavior decoder implements both spatial and global processing pathways.

\begin{figure}[!h]
    \centering
    \includegraphics[scale=0.47]{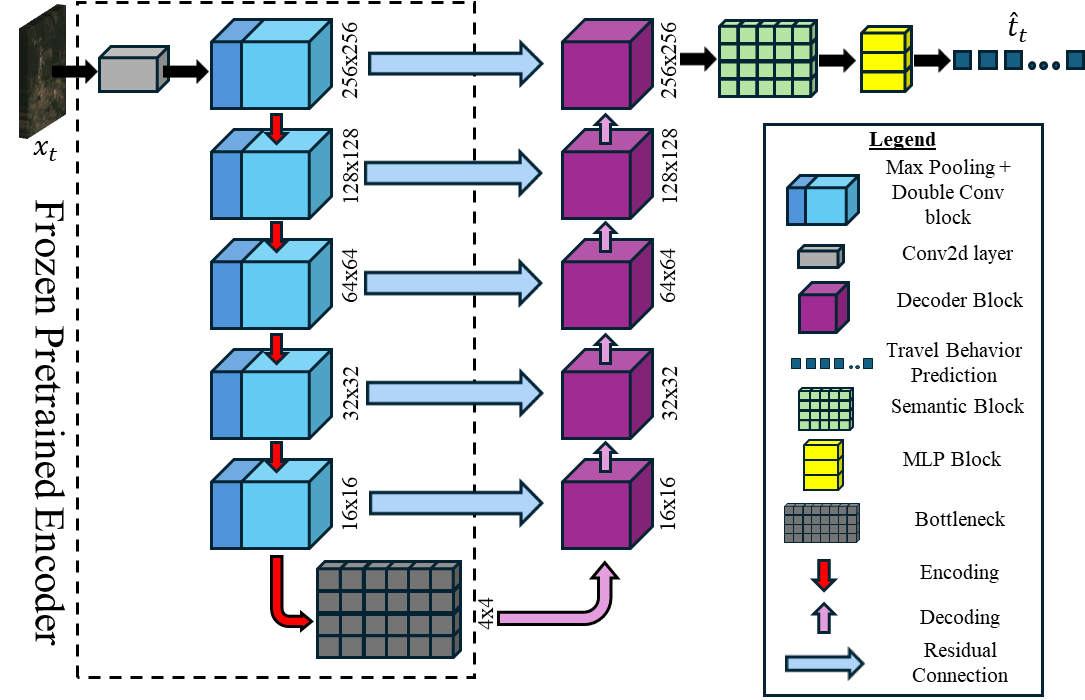}
    \caption{Travel Behavior Prediction Network Architecture, showing the frozen demographic encoder and specialized transportation decoder.}
    \label{fig:tbp-net}
\end{figure}

\subsubsection{Spatial Context Pathway}

The spatial pathway preserves location-specific transportation patterns:

\begin{equation}
S^{(t)}_i = \text{ConvTranspose}_{2 \times 2}(\text{ReLU}(\text{BN}(S^{(t)}_{i-1})))
\end{equation}

for $i \in \{1, 2, \ldots, n_s\}$ where $S^{(t)}_0 = f_e^{frozen}(x_t)$.

\subsubsection{Global Context Pathway}

The global pathway captures region-wide transportation characteristics:

\begin{equation}
\begin{split}
    G^{(t)} = \text{FC}_3(\text{ReLU}(\text{FC}_2(\text{ReLU}(\text{FC}_1\\(\text{AdaptiveAvgPool}(f_e^{frozen}(x_t)))))))
\end{split}
\end{equation}

where $\text{FC}_i$ represents fully connected layers with dimensions $\{512, 256, 128\}$.

\subsubsection{Multi-task Output Formulation}

The travel behavior prediction combines both pathways:

\begin{equation}
\hat{t}_t = [\hat{t}_{mode}, \hat{t}_{vehicles}, \hat{t}_{time}, \hat{t}_{pattern}]
\end{equation}

where each component applies task-specific transformations:

\begin{equation}
\begin{aligned}
\hat{t}_{mode} &= \text{Softmax}(W_{mode}[S^{(t)}_{n_s}, G^{(t)}] + b_{mode}) \\
\hat{t}_{vehicles} &= \text{Softmax}(W_{vehicles}[S^{(t)}_{n_s}, G^{(t)}] + b_{vehicles}) \\
\hat{t}_{time} &= \text{ReLU}(W_{time}[S^{(t)}_{n_s}, G^{(t)}] + b_{time}) \\
\hat{t}_{pattern} &= \sigma(W_{pattern}[S^{(t)}_{n_s}, G^{(t)}] + b_{pattern})
\end{aligned}
\end{equation}

The softmax constraints ensure probability distributions for categorical variables, while ReLU and sigmoid activations handle continuous and binary outputs respectively.

\subsection{Integrated Loss Function}

The complete loss function integrates multiple objectives with mathematical constraints:

\begin{equation}
\mathcal{L}_{total} = \alpha \mathcal{L}_{image} + \beta \mathcal{L}_{demo} + \gamma \mathcal{L}_{travel} + \delta \mathcal{L}_{semantic}
\end{equation}

\subsubsection{Image Prediction Loss}

The spatial prediction loss combines pixel-wise and structural components:

\begin{equation}
\begin{split}
    \mathcal{L}_{image} = \lambda \cdot \frac{1}{HWC}\sum_{i,j,c}(\hat{x}_{t+1}^{(i,j,c)}\\ - x_{t+1}^{(i,j,c)})^2 + (1-\lambda)(1-\text{SSIM}(\hat{x}_{t+1}, x_{t+1}))
\end{split}
\end{equation}

where SSIM is computed as:

\begin{equation}
\text{SSIM}(x, \hat{x}) = \frac{(2\mu_x\mu_{\hat{x}} + C_1)(2\sigma_{x\hat{x}} + C_2)}{(\mu_x^2 + \mu_{\hat{x}}^2 + C_1)(\sigma_x^2 + \sigma_{\hat{x}}^2 + C_2)}
\end{equation}

with $C_1 = (0.01 \times \text{data\_range})^2$ and $C_2 = (0.03 \times \text{data\_range})^2$.

\subsubsection{Demographic and Travel Behavior Losses}

The demographic and travel behavior losses implement L2 regularization:

\begin{equation}
\begin{aligned}
\mathcal{L}_{demo} &= \frac{1}{f}\sum_{j=1}^{f}(\hat{d}_{t+1}^{(j)} - d_{t+1}^{(j)})^2 \\
\mathcal{L}_{travel} &= \frac{1}{m}\sum_{k=1}^{m}(\hat{t}_{t+1}^{(k)} - t_{t+1}^{(k)})^2
\end{aligned}
\end{equation}

\subsubsection{Semantic Consistency Loss}

The semantic loss enforces latent representation alignment:

\begin{equation}
\mathcal{L}_{semantic} = \|\text{Normalize}(S^{(d)}) - \text{Normalize}(B^{(d)})\|_2^2
\end{equation}

where normalization ensures unit variance:

\begin{equation}
\text{Normalize}(x) = \frac{x - \mu_x}{\sigma_x + \epsilon}
\end{equation}

This comprehensive mathematical framework establishes the theoretical foundation for joint spatial-demographic-travel behavior prediction while maintaining computational efficiency and interpretability.

\section{Experiments}

In this section, we outline our experimental setup, including the dataset composition, implementation environment, model configuration, and training procedure. Our experiments aim to validate the effectiveness of the integrated architecture in capturing the complex relationship between spatial patterns and demographic factors for accurate future prediction.

\subsection{Data}

Our architectural framework relies on two complementary data sources: demographic and travel behavior data from the American Community Survey (ACS) and corresponding satellite imagery acquired through Google Earth Engine. This section details the data acquisition process, preprocessing steps, and the final dataset structure used for model training and evaluation.

\subsubsection{Demographic Data}

Demographic and travel behavior data were obtained from the U.S. Census Bureau's American Community Survey (ACS) 1-year estimates, which provide comprehensive data for geographic areas with populations of 65,000 or more. This dataset offers higher temporal resolution compared to the more commonly used 5-year estimates, making it suitable for capturing year-to-year socioeconomic changes. We collected ACS 1-year estimates for the period 2012-2023 (excluding 2020 due to the covid-19 pandemic), extracting 25 key demographic features including population statistics, age distribution, racial composition and educational attainment.

The acquisition process involved using the Census Bureau's API to query and retrieve demographic and travel behavior data. Geographic coordinates for each county were obtained from Census Bureau gazetteer files, providing standardized latitude and longitude values that served as reference points for satellite imagery collection.

To ensure temporal consistency in our analysis, we identified counties that appeared in all study years, resulting in a consistent longitudinal dataset. This constraint was necessary to build reliable temporal prediction models, though it limited our analysis to larger counties due to the population threshold inherent in ACS 1-year estimates. The resulting demographic dataset consists of counties with complete data across all study years, providing a robust foundation for socioeconomic trend analysis.
Table \ref{tab:demographic_features} presents the core demographic features extracted from the ACS, categorized by domain.

\begin{table*}[ht]
\centering
\caption{Demographic features extracted from American Community Survey}
\label{tab:demographic_features}
\begin{tabular}{p{0.15\linewidth} p{0.85\linewidth}}
\hline
\textbf{Feature Category} & \textbf{Variables} \\
\hline
Population & total population, total male, total female, median age \\
\hline
Age Distribution & under 5years, age 5 to 9years, age 10 to 14years, age 15 to 17years, age 18years, age 19years, age 20 to 24years, age 25 to 29years, age 30 to 34years, age 35 to 39years, age 40 to 44years, age 45 to 49years, age 50 to 54years, age 55 to 59years, age 60 to 64years, age 65 to 69years, age 70 to 74years, age 75 to 79years, age 80 to 84years, age 85years and over \\
\hline
Race and Ethnicity & race white alone, race black alone, race american indian alaska native, race asian, race native hawaiian pacific islander, hispanic latino \\
\hline
Education & education total population 25 plus, education bachelors degree, education masters degree, education professional degree, education doctorate degree \\

\hline
Transportation & transportation\_total\_workers, transportation\_drove\_alone, transportation\_carpooled, transportation\_public\_transit\_total, transportation\_taxicab, transportation\_motorcycle, transportation\_bicycle, transportation\_walked, transportation\_other\_means, transportation\_worked\_at\_home, travel\_time\_total, vehicles\_available\_total\_households, vehicles\_available\_no\_vehicle, vehicles\_available\_one\_vehicle, vehicles\_available\_two\_vehicles, vehicles\_available\_three\_plus\_vehicles, departure\_time\_total  \\
\hline
\end{tabular}
\end{table*}

\subsubsection{Satellite Imagery Acquisition}

Corresponding satellite imagery was acquired through Google Earth Engine's cloud-based platform, which provides access to a comprehensive archive of publicly available remote sensing data. For each county in our demographic dataset, we retrieved Landsat 8 Collection 2 Level 2 surface reflectance imagery. The collection process involved several steps to ensure high-quality, representative imagery for each location:

First, a square area of interest (AOI) was defined around each county's centroid coordinates, using a fixed spatial extent of 10 kilometers squared. This standardized approach ensures consistent spatial resolution across all locations while capturing sufficient urban and suburban development patterns. For each AOI, we filtered the Landsat 8 collection to images acquired during the same year as the corresponding demographic data, applying a cloud cover threshold of 20\% to minimize atmospheric interference. In cases where no images met this strict threshold, we employed a relaxed cloud filter of 50\% to ensure data availability.

To account for seasonal variations and potential data gaps, we computed annual median composites for each AOI. This approach creates representative, cloud-free imagery by selecting the median pixel value across all available observations within the year. The pixel-wise median compositing can be formally expressed as:

\begin{equation}
I_{\text{composite}}(x,y,b) = \text{median}\{I_t(x,y,b) : t \in T_y\}
\end{equation}

where $I_{\text{composite}}(x,y,b)$ represents the composite value at pixel location $(x,y)$ for spectral band $b$, $I_t(x,y,b)$ is the surface reflectance value at the same location and band for an image acquired at time $t$, and $T_y$ is the set of all acquisition timestamps within year $y$ for which imagery is available in the filtered collection. This pixel-wise median operation effectively removes transient features such as clouds, shadows, and seasonal anomalies while preserving the persistent landscape characteristics \cite{GORELICK201718}. The resulting composite images were processed to create true-color representations using the red, green, and blue spectral bands, with standardized visualization parameters to ensure consistency across the dataset. Each image was acquired at a resolution of 256×256 pixels, providing sufficient detail to identify urban development patterns, infrastructure, and land cover characteristics.

\begin{figure*}
    \centering
    \includegraphics[scale=0.72]{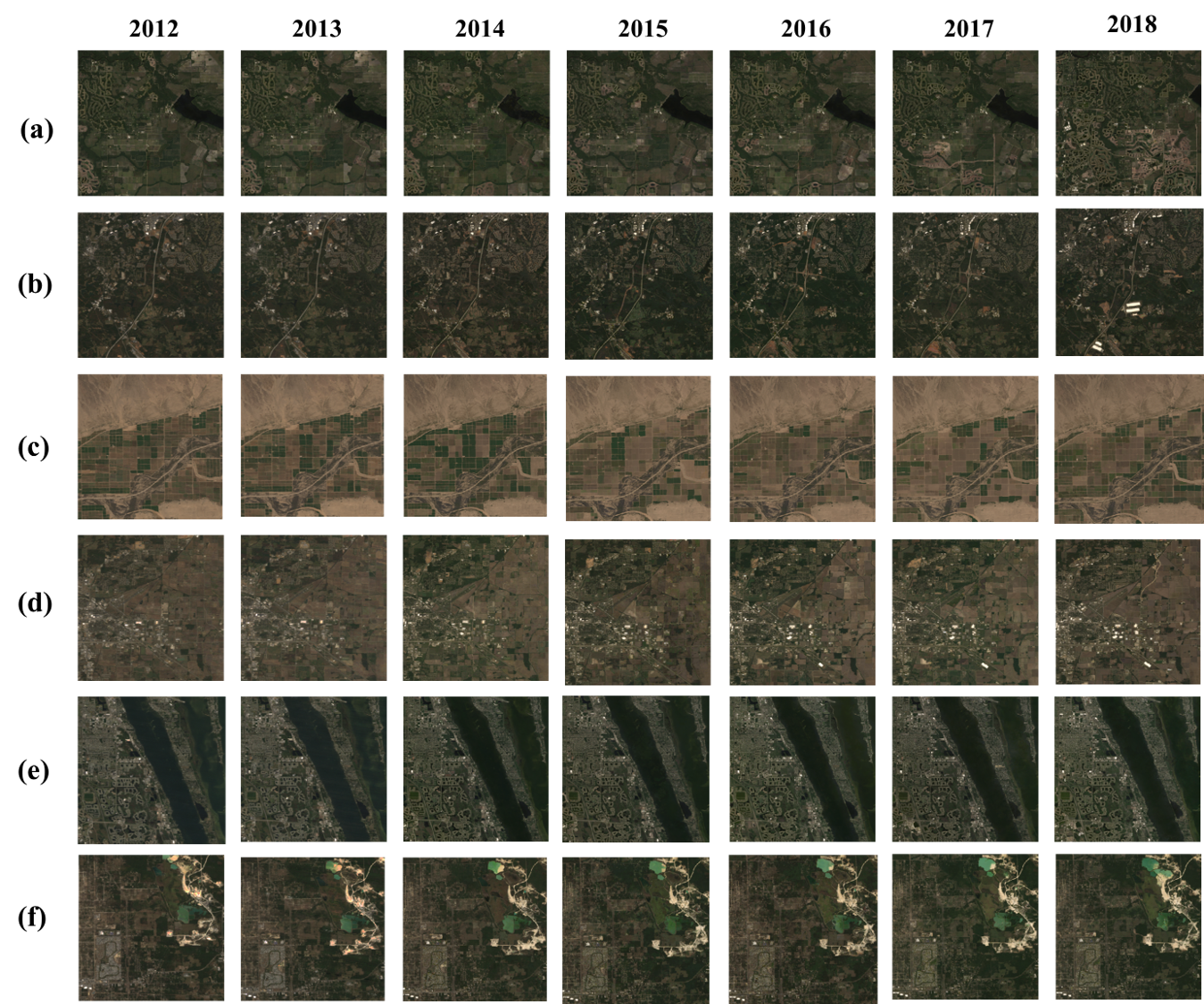}
    \caption{Temporal sequences of sample satellite imagery (2012-2018) demonstrating the subtle, progressive nature of landscape transformations across diverse geographic contexts. Each row (a-f) represents a different county with distinct development patterns: (a) shows gradual suburban transition primarily visible in 2017-2018, (b) illustrates the incremental expansion of transportation infrastructure with increasing connectivity of roads and the emergence of new developed areas (visible as lighter patches) along the main corridor, culminating in noticeable white structures in the bottom right by 2018, (c) displays agricultural land use evolution with minimal structural change, (d) demonstrates dispersed development across a predominantly rural landscape, (e) illustrates stability in established urban-water interfaces with nearly imperceptible change, and (f) shows progressive riparian zone modifications. These examples highlight the challenging nature of the dataset, where year-over-year changes often manifest as subtle shifts in land cover, infrastructure density, and development patterns rather than dramatic transformations.}
    \label{fig:5}
\end{figure*}

\begin{figure*}
    \centering
    \includegraphics[scale=0.5]{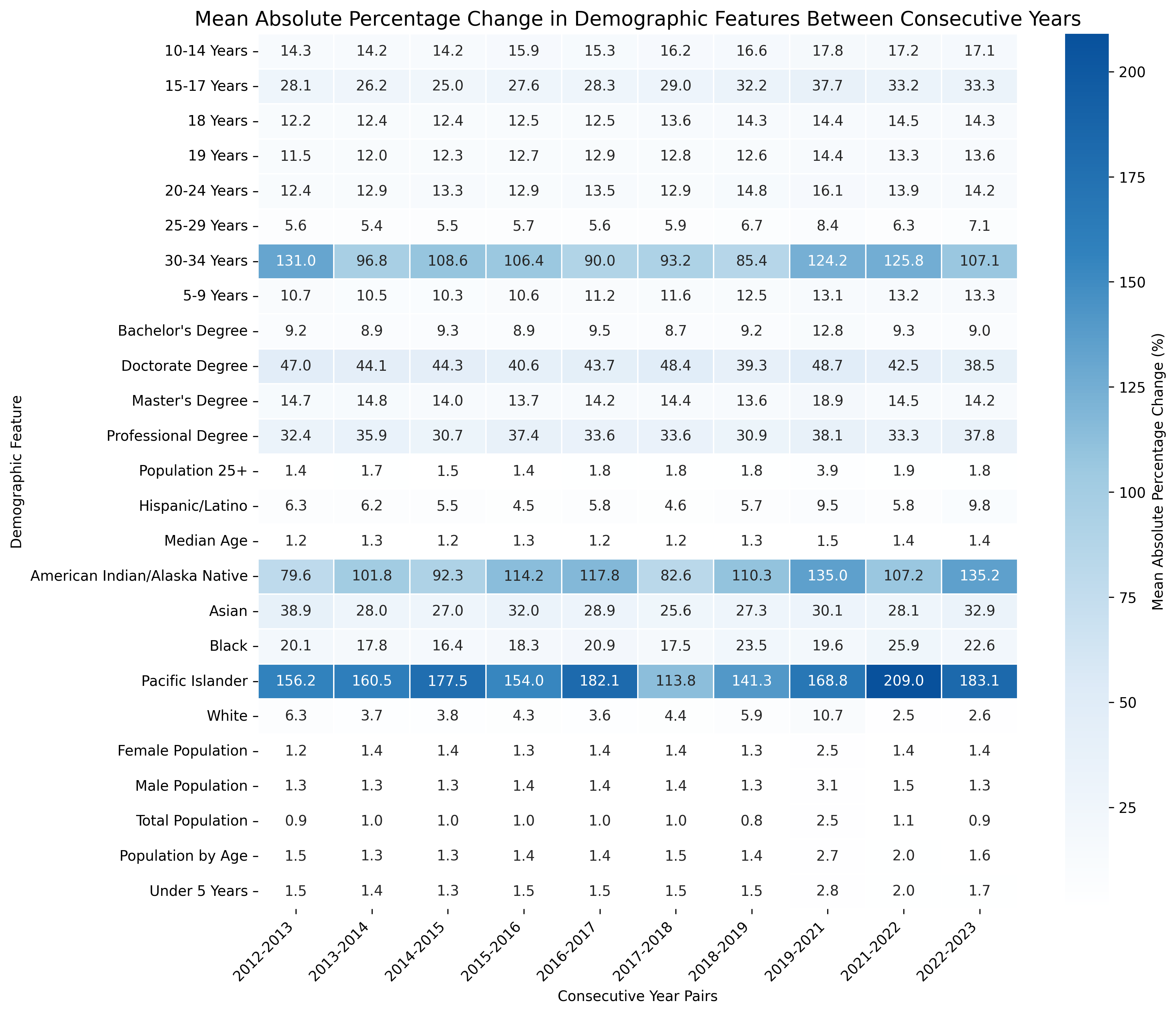}
    \caption{Mean Absolute Percentage Change (MAPC) in demographic features between consecutive years (2012-2023). The heatmap reveals stark variability across different demographic categories, with certain populations exhibiting the most dramatic fluctuations (Pacific Islander: 113-209\%, American Indian/Alaska Native: 79-135\%). Specific age cohorts, particularly 30-34 Years (85-131\%) and 15-17 Years (25-38\%), show substantial year-over-year volatility. Higher education metrics, including Doctorate Degree (39-49\%) and Professional Degree (31-38\%), also demonstrate considerable variability. In contrast, core population metrics (Total Population, Male/Female Population, Median Age) maintain remarkable stability with most changes below 2\%. This pattern of demographic volatility occurring beneath stable overall population trends reveals the complex, spatially heterogeneous nature of demographic shifts that are not immediately apparent in aggregated statistics.}
    \label{fig:demographic_mapc}
\end{figure*}

\textbf{An important characteristic of our dataset is the striking contrast between subtle physical changes and highly variable demographic shifts.} As illustrated in Figure~\ref{fig:demographic_mapc}, demographic variables exhibit substantial heterogeneity in their year-over-year percentage changes, with some features demonstrating extreme volatility. Several racial and ethnic categories show significant fluctuations, with Pacific Islander populations experiencing mean absolute percentage changes (MAPC) of 113-209\% between consecutive years, while American Indian/Alaska Native populations show variations of 79-135\%. Similarly, certain age cohorts (particularly 30-34 Years at 85-131\%) and advanced education metrics (Doctorate Degree at 38-49\%) exhibit considerable variability. Conversely, Figure~\ref{fig:5} demonstrates that these demographic shifts occur alongside nearly imperceptible changes in the physical landscape, where satellite imagery reveals only subtle, incremental modifications to land use and infrastructure across multiple years.


\subsubsection{Data Preprocessing and Integration}

The preprocessing pipeline addressed several challenges inherent in both demographic and satellite data sources. For demographic variables, we implemented a multi-stage cleaning process that involved:

\begin{equation}
x_{norm} = \frac{x - \mu_x}{\sigma_x}
\end{equation}

where $x$ represents the actual demographic feature value, and $\mu_x$ and $\sigma_x$ are the mean and standard deviation of the feature across all counties and years. This standardization was essential for normalizing the widely varying scales of demographic features, from population counts in the millions to percentage-based metrics.

Missing values, typically represented as negative codes (-666666666 or -999999999) in Census data, were identified and replaced with null values. We then applied outlier detection using the interquartile range method:

\begin{equation}
\text{Outlier if } x < Q_1 - 1.5 \times IQR \text{ or } x > Q_3 + 1.5 \times IQR
\end{equation}

where $Q_1$ and $Q_3$ represent the first and third quartiles, and $IQR = Q_3 - Q_1$. Identified outliers were examined individually and either corrected or excluded based on domain knowledge and data context.

For satellite imagery, preprocessing focused on ensuring radiometric consistency and spatial standardization. Atmospheric correction had already been applied in the Landsat 8 Collection 2 Level 2 products, providing surface reflectance values. Our additional preprocessing steps included:

\begin{equation}
I_{scaled} = \gamma \times \text{clip}(I, v_{min}, v_{max})
\end{equation}

where $I$ represents the original surface reflectance values, $v_{min}$ and $v_{max}$ are the scaling boundaries (7000 and 30000 for Landsat 8 surface reflectance), and $\gamma = 1.4$ is a gamma correction factor applied to enhance visual contrast. The resulting images were standardized for consistent interpretation across different geographic contexts and lighting conditions.

The final integrated dataset links each county's demographic profile with its corresponding satellite imagery for each year, creating spatiotemporal data pairs suitable for predictive modeling. This integration enables the exploration of bidirectional relationships between physical infrastructure (visible in satellite imagery) and socioeconomic characteristics (captured in demographic data). The dataset consists of 825 corresponding demographic, travel behavior and satellite images for every year from 2012 to 2023 (excluding 2020), making the total number of datapoints 9075.

\subsection{Implementation Environment}
The implementation process involved the use of high-performance computing resources to handle the computationally intensive nature of training deep neural networks on spatio-temporal data. The experimental environment specifications are detailed in Table \ref{tab:environment}.

\begin{table}[ht]
\centering
\caption{Implementation environment specifications}
\label{tab:environment}
\begin{tabular}{ll}
\hline
\textbf{Component} & \textbf{Specification} \\
\hline
Hardware & Nvidia A100 GPU (40GB VRAM) \\
System memory & 64GB RAM \\
Programming language & Python 3.8 \\
Deep learning framework & PyTorch 1.9 \\
Development environment & Visual Studio Code \\
Additional libraries & NumPy, Pandas, PIL, Matplotlib \\
\hline
\end{tabular}
\end{table}

The implementation was developed using PyTorch due to its flexibility in designing complex neural network architectures and its efficient utilization of GPU acceleration.

\subsection{Model Configuration}
The integrated model architecture was configured with carefully selected hyperparameters to balance model capacity, training efficiency, and generalization capability. The hyperparameter configuration is summarized in Table \ref{tab:hyperparams}.

\begin{table}[ht]
\centering
\caption{Model hyperparameters and configuration}
\label{tab:hyperparams}
\begin{tabular}{ll}
\hline
\textbf{Hyperparameter} & \textbf{Value} \\
\hline
\multicolumn{2}{l}{\textit{Training Parameters}} \\
\hline
Batch size & 8 \\
Learning rate & $3 \times 10^{-4}$ \\
Number of epochs & 200 \\
Optimizer & Adam \\
Loss weighting ($\alpha$) & 0.7 (image), 0.3 (demographic) \\
Image loss weighting ($\gamma$) & 0.7 (MSE), 0.3 (SSIM) \\
\hline
\end{tabular}
\end{table}

\subsubsection{Data Evaluation and Validation}
To ensure data quality, we implemented a multi-stage validation process for both the satellite imagery and sociodemographic components of our dataset. For the satellite imagery collection, we conducted a manual verification process by randomly sampling approximately 50 Areas of Interest (AOIs) and cross-referencing their geographic coordinates with high-resolution imagery from Google Maps. This verification process confirmed that our acquisition methodology produced imagery that accurately represented the intended geographic regions and captured relevant development patterns. The comparison involved examining key landscape features such as road networks, water bodies, and building clusters to verify proper geolocation and registration of the extracted imagery.

For demographic and travel behavior data, our evaluation strategy focused on internal consistency and temporal coherence. We implemented a series of validation checks including:

\begin{itemize}   
    \item \textbf{Internal coherence validation}: We verified relational consistency between logically connected variables (ensuring that subcategories of transportation modes summed to match reported totals) and identified any mathematical inconsistencies for correction.
    
    \item \textbf{Completeness assessment}: We evaluated the percentage of missing values across years and variables to ensure sufficient data density for robust modeling. Counties with missing values across key variables were excluded from the final dataset.
\end{itemize}

This validation approach, provided assurance of data quality through examination of internal consistency and alignment with expected patterns of demographic change. The satellite imagery validation further confirmed the spatial accuracy of our dataset, establishing a reliable foundation for the integrated spatiotemporal modeling approach employed in our research.

\section{Evaluation Metrics}
To comprehensively evaluate the proposed architecture, the study employs a diverse set of metrics addressing both the spatial, demographic and travel behavior prediction aspects. This section outlines our evaluation methodology and presents a comparative analysis with leading architectures in spatio-temporal prediction.

\subsection{Performance Metrics}

Our evaluation framework incorporates complementary metrics for each prediction task. For spatial imagery assessment, we utilize Mean Squared Error (MSE) for pixel-level accuracy, Structural Similarity Index Measure (SSIM) for perceptual quality, and Peak Signal-to-Noise Ratio (PSNR) for image fidelity. For demographic and travel behavior predictions, we employ MSE and coefficient of determination (R$^2$) to assess prediction accuracy across different variables after normalization to ensure fair comparison across diverse scales.

\subsection{Multi-Horizon Error Analysis}

We conduct comprehensive error analysis across 1-year, 2-year, and 3-year prediction horizons to evaluate how model performance degrades with increasing temporal distance. For each horizon $h \in \{1, 2, 3\}$ years, we compute:

\textbf{Image prediction errors}: For each pixel $(i,j)$ in predicted image $\hat{I}_{t+h}$ and actual image $I_{t+h}$:
\begin{equation}
e^{img}_{t+h}(i,j,c) = \hat{I}_{t+h}(i,j,c) - I_{t+h}(i,j,c)
\end{equation}

We analyze error distributions using quantile-quantile (QQ) plots to assess normality and identify systematic biases. This approach reveals whether errors accumulate linearly or non-linearly with increasing prediction horizons, providing insights into long-term forecasting reliability.

\subsection{Change Heatmap Analysis}

Changes in sequential satellite imagery are often subtle and difficult to identify through visual inspection alone. Many landscape transformations occur gradually over time, with incremental modifications that may escape detection without computational assistance. To address this challenge, we have implemented a comprehensive change heatmap analysis methodology that quantifies and visualizes the differences between temporal sequences of satellite imagery, allowing for precise identification of areas undergoing transformation.

For a sequence of historical satellite images spanning multiple years, we denote the image from year $t$ as $I_t$. Given a base year $t_0$ (in our implementation, $t_0 = 2012$) and a set of subsequent years $\{t_1, t_2, ..., t_n\}$, we compute change heatmaps that highlight the spatial differences between the base year and each subsequent year.

For each pixel position $(i,j)$ in the image domain $\Omega$, the change heatmap $H_{t_0 \rightarrow t_k}$ between the base year $t_0$ and a subsequent year $t_k$ is computed as:

\begin{equation}
H_{t_0 \rightarrow t_k}(i,j) = \log(1 + \sum_{c \in \{R,G,B\}} |I_{t_k}(i,j,c) - I_{t_0}(i,j,c)|)
\end{equation}

where $c$ represents the color channels (red, green, blue). The logarithmic transformation enhances the visibility of subtle changes while preventing extreme values from dominating the visualization.

\subsubsection{Computation of Target and Forecasted Change Heatmaps}

In the context of our predictive modeling framework, we compute two types of change heatmaps:

\textbf{Target Change Heatmap:} This represents the actual changes that occurred between the base year $t_0$ and target year $t_k$. For each pixel $(i,j)$:

\begin{equation}
H^{target}_{t_0 \rightarrow t_k}(i,j) = \log(1 + \sum_{c \in \{R,G,B\}} |I^{actual}_{t_k}(i,j,c) - I_{t_0}(i,j,c)|)
\end{equation}

where $I^{actual}_{t_k}$ is the actual observed satellite image at year $t_k$.

\textbf{Forecasted Change Heatmap:} This represents the predicted changes between the base year $t_0$ and the model's forecast for year $t_k$. For each pixel $(i,j)$:

\begin{equation}
H^{forecast}_{t_0 \rightarrow t_k}(i,j) = \log(1 + \sum_{c \in \{R,G,B\}} |\hat{I}_{t_k}(i,j,c) - I_{t_0}(i,j,c)|)
\end{equation}

where $\hat{I}_{t_k}$ is the model's forecasted satellite image for year $t_k$.

The comparison between these two heatmaps provides critical insights into the model's ability to accurately predict not just the future state of the landscape but also to correctly identify the specific locations and magnitudes of change.

\subsubsection{Multi-temporal Change Accumulation}

To capture changes that persistently occur across multiple years, we implement an accumulation mechanism that aggregates changes over time. For a sequence of years $\{t_1, t_2, ..., t_n\}$, the accumulated change frequency map $F$ is computed as:

\begin{equation}
F(i,j) = \frac{1}{n} \sum_{k=1}^{n} \mathbf{1}[H_{t_0 \rightarrow t_k}(i,j) > \tau]
\end{equation}

where $\mathbf{1}[\cdot]$ is the indicator function, and $\tau$ is a threshold that defines significant change. This frequency map highlights areas that consistently undergo transformation across multiple years, providing insight into persistent rather than transient changes.

\subsubsection{Visualization and Interpretation}

\begin{figure*}[!h]
\centering
\includegraphics[scale=0.74]{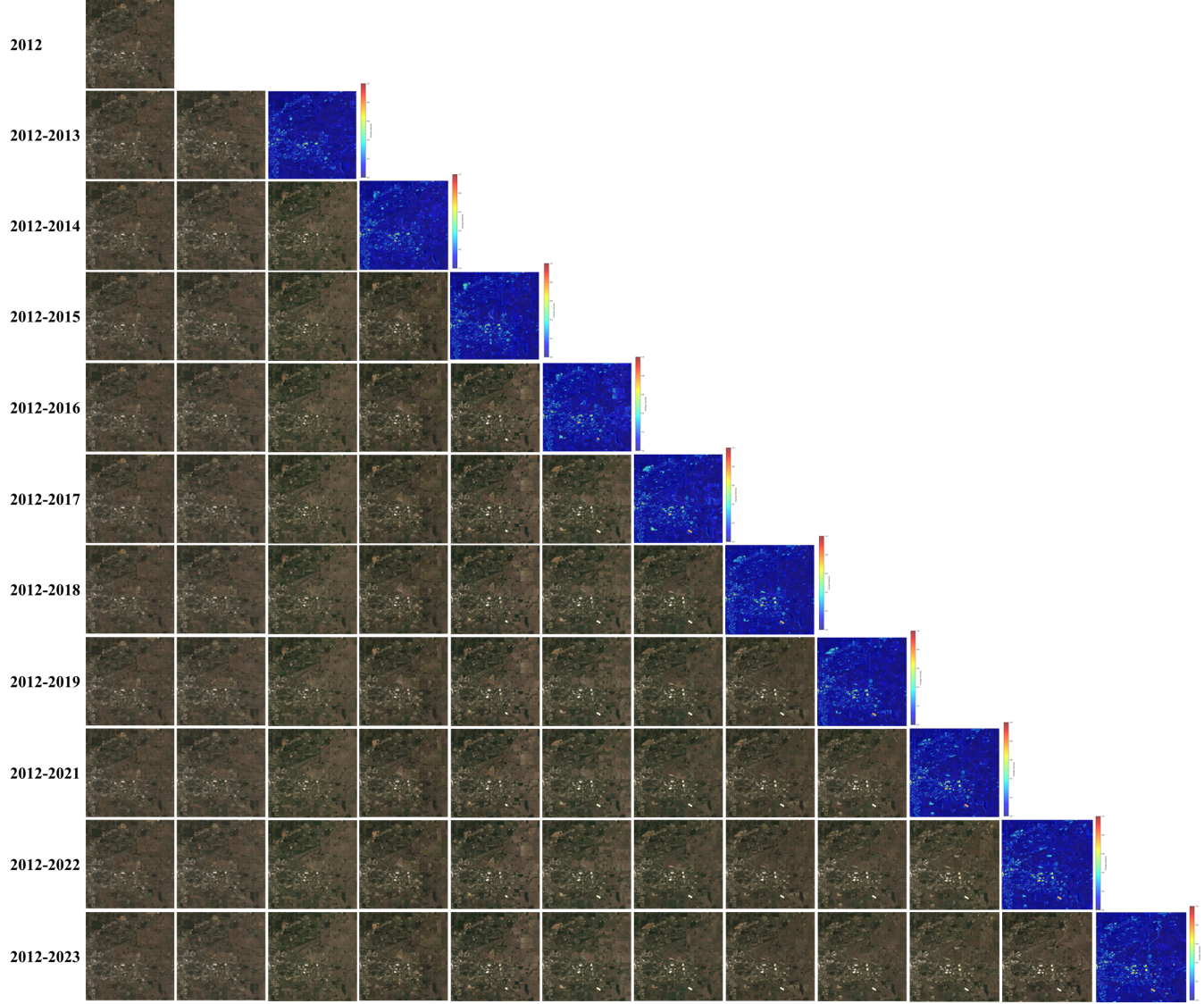}
\caption{\textbf{Progressive Change Detection in Satellite Imagery (2012-2023).} The figure illustrates the detection of subtle changes in satellite imagery over an 11-year period. The top row shows the base image from 2012. Each subsequent row represents a comparison between 2012 and a later year, with the rightmost column displaying the computed change heatmap. The blue-yellow-red color gradient in the heatmaps indicates increasing magnitude of change. Note how certain areas consistently show higher change intensity across multiple years, indicating persistent landscape transformation. While changes in the original imagery may appear subtle to the human eye, the heatmaps effectively highlight and quantify areas undergoing significant transformation.}
\label{fig:change_heatmap_sequence}
\end{figure*}

Figure \ref{fig:change_heatmap_sequence} presents a comprehensive visualization of the change detection process across an 11-year period (2012-2023). The top row displays the base satellite image from 2012. Each subsequent row shows a progressive comparison between 2012 and a later year, with the corresponding change heatmap displayed at the end of each row. The heatmaps uses a blue-yellow-red colormap, where blue indicates minimal change, yellow represents moderate change, and red signifies substantial transformation.

While the changes in the raw satellite imagery are subtle and challenging to discern through visual inspection alone, the heatmaps effectively highlight areas of transformation. Particularly notable is the ability of these heatmaps to reveal patterns that may indicate urban development, agricultural transitions, or environmental changes that occur incrementally over time.

The progressive nature of the visualization, illustrated in figure \ref{fig:change_heatmap_sequence} demonstrates how some areas experience consistent change (appearing in similar locations across multiple years' heatmaps), while others undergo more sudden or periodic transformations. This temporal dimension of change detection provides valuable context for understanding landscape dynamics.

Our approach enables the detection of incremental changes that might otherwise go unnoticed through visual inspection. By providing a quantitative foundation for change analysis, the methodology supports evidence-based decision-making and enhances our understanding of complex spatiotemporal processes that shape our landscapes.

\subsection{Ablation Studies}

To understand the contribution of individual architectural components to the overall performance of the integrated model, the study also conducted a series of ablation studies. This approach involves removing or modifying specific components of the architecture and evaluating the resulting model's performance. Table \ref{tab:ablation} presents the different model variants examined in our ablation studies.

\begin{table*}[ht]
\centering
\caption{Ablation model variants}
\label{tab:ablation}
\begin{tabular}{c p{0.25\linewidth} c c c}
\hline
\textbf{Model} & \textbf{Description} & \textbf{Encoder Type} & \textbf{Gated Skip} & \textbf{Demographic} \\
 &  & \textbf{(Dense/2DConv)} & \textbf{Connection} & \textbf{Predictor} \\
\hline
Model 1 (Baseline Model) & Without demographic predictor and gated skip connections & 2DConv  & \ding{55} & \ding{55} \\
\hline
Model 2 & Without DenseNet blocks & 2DConv & \checkmark & \checkmark \\
\hline
Model 3 & Without gated skip connections & Dense & \ding{55} & \checkmark \\
\hline
Model 4 & Without demographic predictor & Dense & \checkmark & \ding{55} \\
\hline
Model 5 (Ours) & Full Integrated model & Dense & \checkmark & \checkmark \\

\hline
\end{tabular}
\end{table*}

These ablation studies provide insights into which architectural components contribute most significantly to the model's performance. By comparing the performance metrics across these model variants, we can identify the relative importance of:

\begin{enumerate}
    \item Dense connectivity patterns in feature extraction
    \item Gated mechanisms in skip connections
    \item Demographic verification through the frozen predictor
\end{enumerate}

The results of these ablation studies not only validate our architectural choices but also provide guidance for future refinements and adaptations of the model to different domains and applications.

\subsection{Comparative Evaluation with State-of-the-Art Models}

We compare against two state-of-the-art spatio-temporal prediction models adapted for multimodal processing:

\textbf{Predictive Recurrent Neural Network V2 (PredRNNV2):} This model introduces Causal Spatio-Temporal Long Short-Term Memory (CST-LSTM) cells that maintain separate memory states for spatial and temporal information. The CST-LSTM architecture enables effective capture of both immediate spatial context and longer-term temporal dynamics through its dual-memory mechanism. We extended the original architecture by integrating a demographic processing component that encodes demographic features through a multi-layer perceptron (MLP), reshapes them to match spatial dimensions, and fuses them with visual features via dedicated fusion layers.

\textbf{Eidetic 3D Long Short-Term Memory (E3DLSTM):} This model implements a persistent memory mechanism designed for long-term spatio-temporal modeling through eidetic memory components that retain information from extended time sequences. The architecture employs 3D convolutions to process spatial and temporal dimensions simultaneously, enabling direct modeling of complex spatio-temporal patterns. We adapted E3DLSTM by incorporating a sequential demographic encoder that processes time series demographic features and a spatial projection mechanism that transforms encoded features to match visual feature dimensions before fusion.

The comparison focuses on Structural Similarity Index Measure (SSIM) for visual quality assessment and demographic prediction loss (Demo-loss) for evaluating demographic consistency. This evaluation reveals the effectiveness of our architectural design choices in maintaining both spatial accuracy and sociodemographic plausibility compared to existing approaches that primarily optimize for visual continuity.

\section{Results and Discussion}
\subsection{Demographic Predictor Performance}

\begin{figure*}
    \centering
    \includegraphics[scale=0.68]{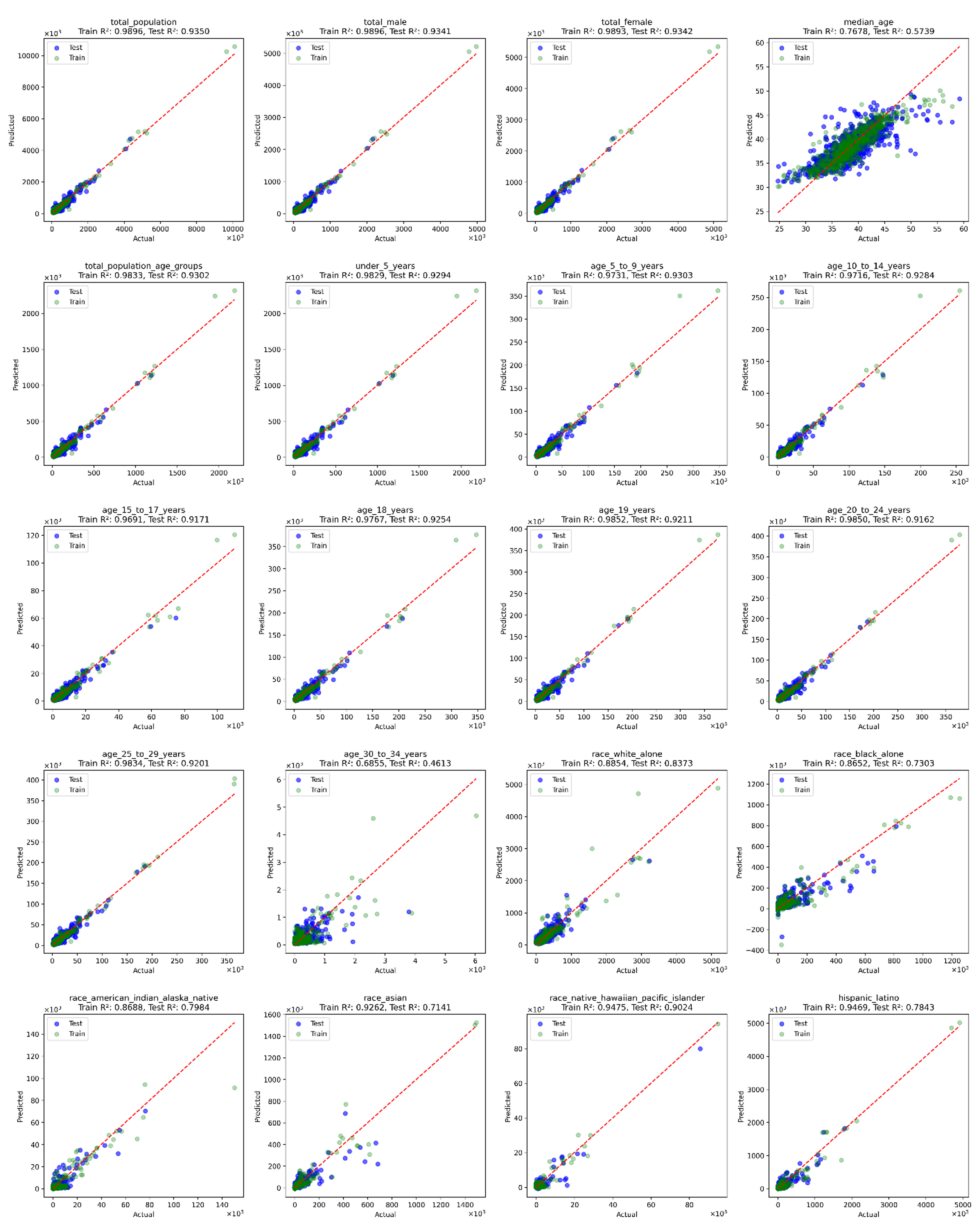}
    \caption{Scatter plots of predicted versus actual values for population, sex, age distributions, racial and ethnic demographics. The diagonal red line represents perfect prediction. The green dots represent plots of predicted versus actual values during training, while the blue dots are for the predicted versus actual values during testing.}
    \label{fig:4.1}
\end{figure*}

\begin{figure*}
    \centering
    \includegraphics[scale=0.68]{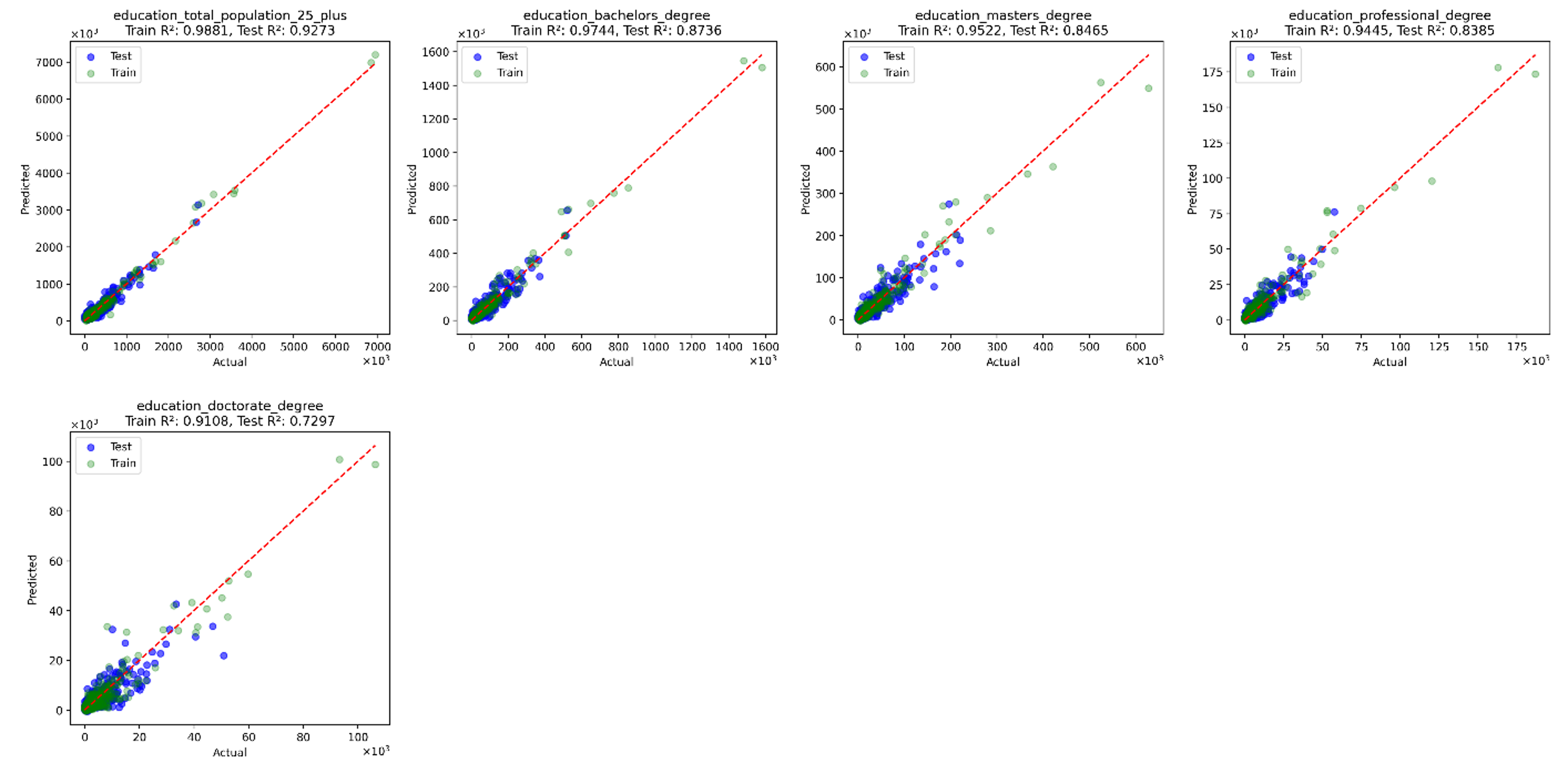}
    \caption{Scatter plots of predicted versus actual values for educational attainment features. The diagonal red line represents perfect prediction. The green dots represent plots of predicted versus actual values during training, while the blue dots are for the predicted versus actual values during testing.}
    \label{fig:4.2}
\end{figure*}

As shown in figures \ref{fig:4.1} and \ref{fig:4.2}, the demographic predictor demonstrates robust performance with an overall R$^2$ of 0.799 and Mean Squared Error (MSE) of 0.145, successfully capturing approximately 80\% of demographic variance from satellite imagery. This finding emphasizes the bidirectional relationship that exists between built environment characteristics and sociodemographic patterns \cite{cervero1997travel}.

Fundamental population metrics achieve exceptional accuracy (total population R$^2$: 0.935, gender-specific predictions R$^2$: 0.934), supporting theories of residential sorting where population density patterns reflect systematic spatial organization \cite{mieszkowski1993causes}. Educational attainment predictions reveal a critical spatial hierarchy: bachelor's degrees achieve R$^2$ of 0.874 while doctorate predictions show substantially lower accuracy (R$^2$: 0.730). This performance differential directly supports Florida's findings that highly educated populations exhibit distinct geographic clustering patterns \cite{florida2011creative}. Florida demonstrated that creative class workers (including those with advanced degrees) concentrate in specific metropolitan areas with particular amenities and institutional infrastructure, creating highly uneven spatial distributions. Our lower prediction accuracy for doctorate holders suggests their residential patterns are less predictable from general built environment features visible in satellite imagery, likely because their location decisions are driven by highly specialized institutional factors including universities, research centers, and knowledge economy hubs—that create concentrated but geographically irregular clusters. In contrast, bachelor's degree holders show more predictable spatial patterns, consistent with broader urban development processes and housing market dynamics that leave clearer signatures in the physical landscape.

Racial and ethnic demographic predictions reveal systematic accuracy differences that directly reflect the spatial imprint of historical segregation processes. White population predictions achieve higher accuracy (R$^2$: 0.837) compared to Black population predictions (R$^2$: 0.730), a pattern that aligns with Massey and Denton's seminal analysis of American residential segregation \cite{massey2019american}. Their research demonstrated that Black populations remain highly segregated in spatially concentrated, often physically distinct neighborhoods with clear boundaries—patterns that would theoretically be more visible in satellite imagery and thus more predictable. However, our lower prediction accuracy for Black populations suggests these segregated areas may have more complex internal spatial organization.

\subsection{Multi-Horizon Error Analysis}
\begin{figure}[ht]
\centering
\includegraphics[scale=0.25]{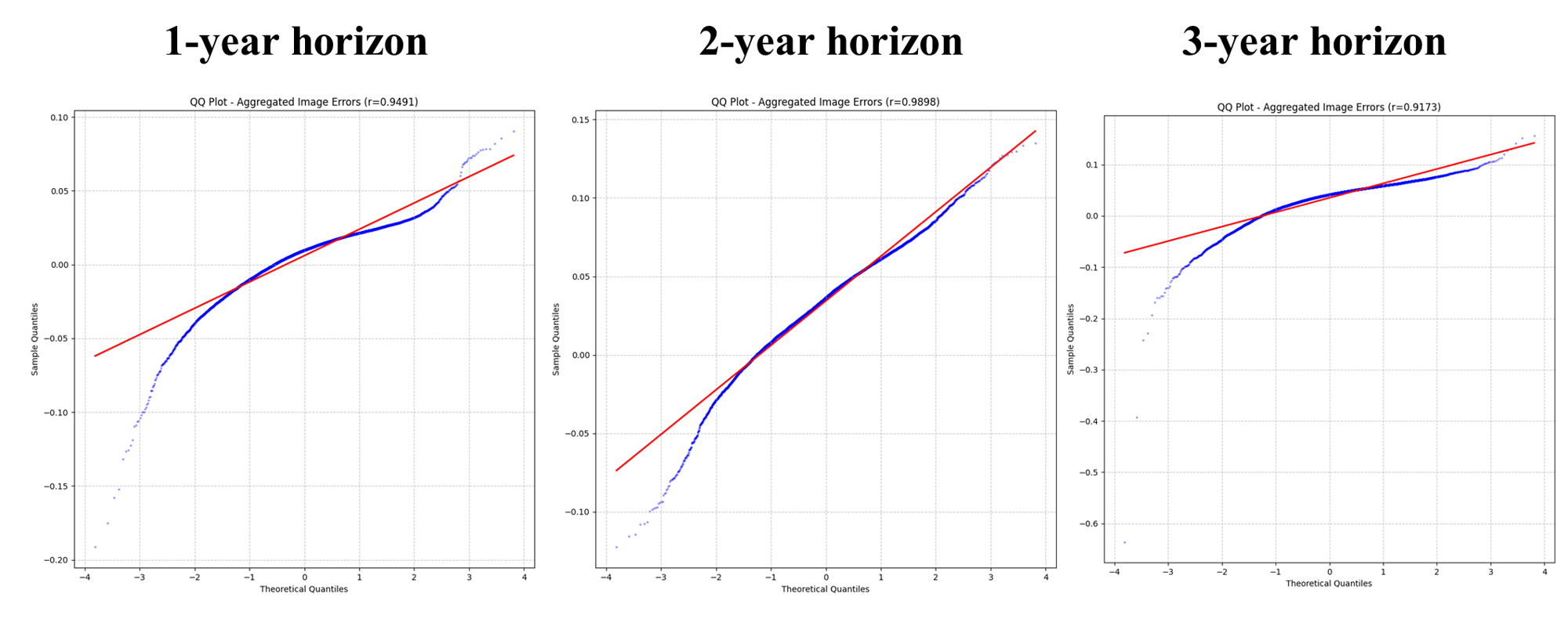}
\caption{QQ plots of aggregated image prediction errors for 1-year, 2-year, and 3-year horizons. Blue points represent observed error quantiles plotted against theoretical quantiles from a normal distribution. Red lines represent perfect normality. The correlation coefficient ($r$) indicates the degree to which errors follow a normal distribution.}
\label{fig:img_qq_plots}
\end{figure}

The multi-horizon error analysis demonstrates that the proposed model maintains consistent prediction accuracy across different temporal scales. As shown in Figure \ref{fig:img_qq_plots}, the quantile-quantile (QQ) plots reveal that image prediction errors closely follow normal distributions, with strong correlation coefficients of 0.9491, 0.9898, and 0.9173 for 1-, 2-, and 3-year prediction horizons, respectively. These high correlation values indicate that prediction errors are systematic rather than random, suggesting reliable model performance across extended time periods. The observed temporal stability in error patterns provides empirical support for path-dependency theories in urban development. According to these theories, current spatial configurations create constraints and opportunities that strongly influence future development trajectories \cite{arthur1994increasing}. The consistent prediction accuracy across multiple years suggests that urban landscapes follow predictable evolutionary patterns, where existing infrastructure, land use arrangements, and demographic distributions establish pathways that guide subsequent changes in measurable and predictable ways.

\subsection{Ablation Study Results}

The ablation analysis, summarized in table \ref{tab:ablation_results} validates architectural components' contributions, with the integrated model achieving superior performance (MSE: 0.00379, SSIM: 0.8342) compared to variants lacking key components.

\begin{table}[ht]
\centering
\caption{Ablation study results across model variants}
\label{tab:ablation_results}
{\small
\begin{tabular}{c c c c}
\hline
\textbf{Model \#} & \textbf{MSE} ($\downarrow$) & \textbf{SSIM} ($\uparrow$) & \textbf{PSNR} ($\uparrow$) \\
\hline
Model 1 (Baseline)   & 0.00425 & 0.8211           & 25.40             \\
Model 2             & 0.00393 & 0.8291           & 25.60             \\
Model 3             & 0.00386 & 0.8265           & \textbf{25.77}    \\
Model 4             & 0.00391 & 0.8311           & 25.63             \\
Model 5 (ours)      & \textbf{0.00379} & \textbf{0.8342} & 25.17         \\
\hline
\end{tabular}
}
\end{table}

The DenseNet encoder provides the largest performance improvement, reducing MSE by 3.7\% through effective feature reuse across temporal sequences. Gated residual connections contribute a 1.8\% MSE reduction by enabling selective feature transmission during spatial reconstruction. The demographic predictor adds a 3.2\% MSE improvement, demonstrating that adding sociodemographic context enhances spatial prediction accuracy. The baseline model shows 12.1\% higher MSE compared to the full integrated architecture, showing that these components work better together than individually.
Despite the fact that our proposed model (Model 5) attains the best MSE and SSIM, its PSNR is marginally lower than the baseline (25.17 dB against 25.40 dB). This slight reduction in PSNR reflects a minor increase in high‐frequency reconstruction error, which is outweighed by the substantial gains in structural similarity and demographic fidelity. In practice, the improved perceptual quality, which is captured by SSIM and the lower MSE deliver more reliable predictions for urban planning applications, despite the small PSNR trade‐off.

\subsection{State-of-the-Art (SOTA) Model Comparison}

The comparative evaluation reveals significant performance differences between our integrated approach and adapted versions of SOTA spatio-temporal prediction models, particularly in demographic consistency. This is shown in table \ref{tab:model_comparison}.

\begin{table}[ht]
\centering
\caption{Performance comparison of different model architectures}
\label{tab:model_comparison}
\begin{tabular}{l c c}
\hline
\textbf{Model} & \textbf{SSIM} $\uparrow$ & \textbf{Demo-loss} $\downarrow$ \\
\hline
Ours & 0.8342 & 0.14 \\
PredRNN-V2 & 0.8218 & 0.95 \\
E3DLSTM & 0.8146 & 0.96 \\
\hline
\end{tabular}
\end{table}

\textbf{Visual Quality Performance:} Our model achieves the highest SSIM score (0.8342), representing a 1.5\% improvement over PredRNN-V2 (0.8218) and a 2.4\% improvement over E3DLSTM (0.8146). While these differences appear modest, they represent meaningful improvements in structural similarity for satellite imagery prediction. The superior performance demonstrates that our proposed demographics-informed neural network produces images that better preserve spatial relationships and visual coherence compared to the recurrent architectures of the baseline models.

\textbf{Demographic Consistency Analysis:} The most striking difference lies in demographic consistency, where our model significantly outperforms both baselines. Our demographic loss of 0.14 represents an 85\% improvement over PredRNN-V2 (0.95) and an 86\% improvement over E3DLSTM (0.96). This gap reveals a key limitation in existing spatio-temporal prediction approaches: despite our extensive modifications to incorporate demographic data processing, these models fundamentally struggle to establish strong demographic correlations in their predictions despite their visual quality performance.

\subsection{Travel Behavior Predictor Performance}

\begin{figure*}
    \centering
    \includegraphics[scale=0.32]{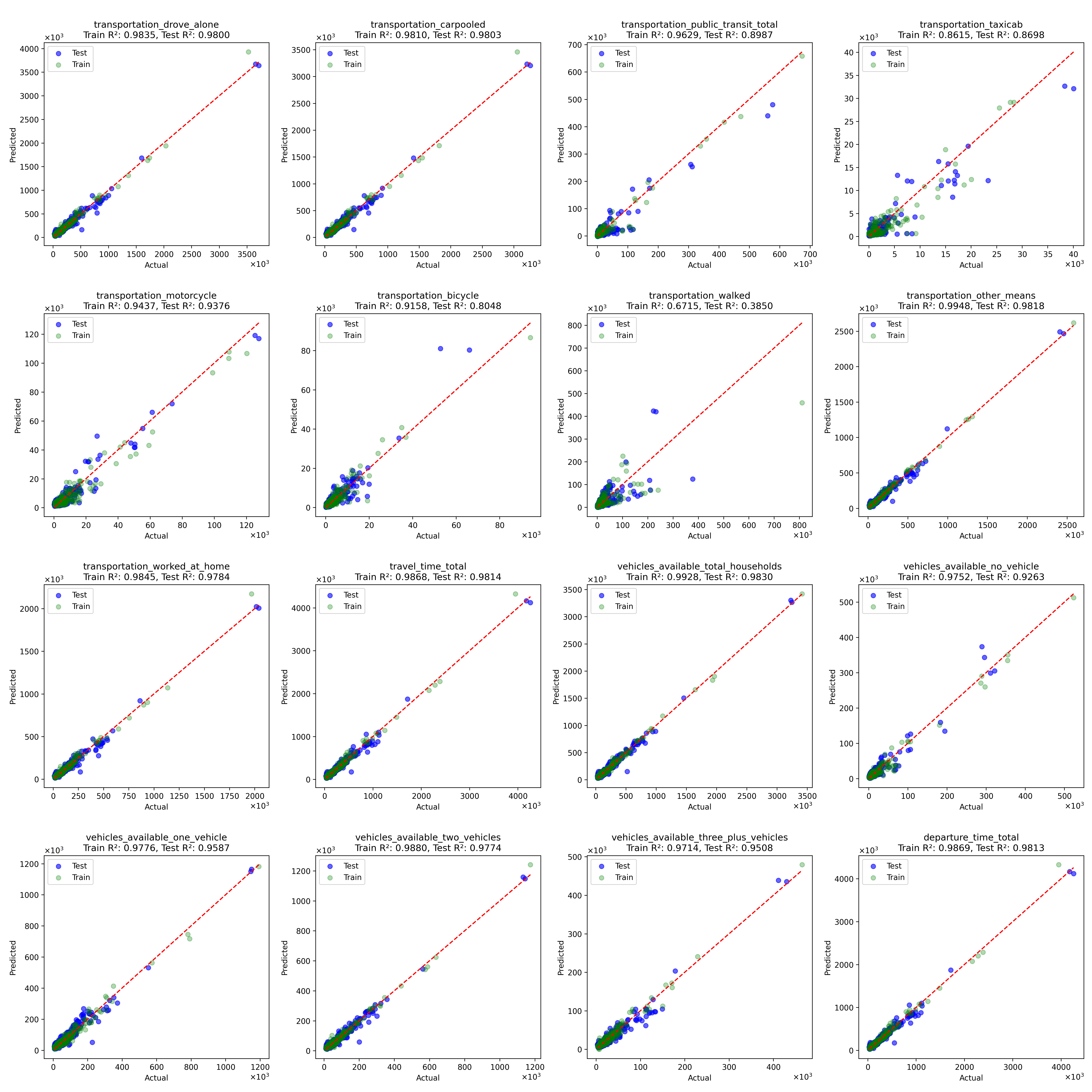}
    \caption{Scatter plots of predicted versus actual values for transportation modes, travel times, and vehicle availability metrics. The diagonal red line represents perfect prediction. Blue dots represent test predictions while green dots indicate training predictions. Each subplot shows the corresponding R$^2$ values for both training and testing sets.}
    \label{fig:travel_behavior}
\end{figure*}

The Travel Behavior Predictor achieves strong performance (R$^2$: 0.911, MSE: 0.078), supporting established theories of built environment-travel behavior relationships \cite{ewing2001travel}. High accuracy in temporal patterns (departure time R$^2$: 0.981, travel time R$^2$: 0.981) reflects the strong spatial embedding of activity-travel patterns, consistent with time geography principles \cite{hagerstrand1970people}. Figure \ref{fig:travel_behavior} shows the scatter plots of predicted versus actual values for transportation modes, travel times, and vehicle availability metrics.

Private vehicle metrics demonstrate exceptional accuracy (drove alone R$^2$: 0.980), directly confirming Newman and Kenworthy's analysis of gasoline consumption and urban structure \cite{newman1989gasoline}. Their comparative study of global cities found that automobile dependence correlates strongly with specific built environment characteristics: low density development, extensive road networks, abundant parking infrastructure, and separated land uses. Our high prediction accuracy for private vehicle usage indicates these physical features create distinctive spatial signatures visible in satellite imagery. The strong predictability suggests that automobile-oriented development produces consistent, recognizable patterns—wide roads, large parking areas, low-density housing, strip commercial development—that our model can reliably identify and associate with high rates of driving alone. This finding validates Newman and Kenworthy's argument that urban form fundamentally shapes transportation behavior, as the physical infrastructure supporting automobile dependence becomes spatially embedded and detectable through remote sensing. Variable performance across transportation modes (walking R$^2$: 0.385, cycling R$^2$: 0.805) supports theories of modal hierarchy in spatial analysis, where infrastructure-intensive modes exhibit stronger spatial signatures than flexible modes dependent on micro-scale environmental factors \cite{saelens2003environmental}.

\subsection{Theoretical and Practical Implications}

These findings advance several key debates in spatial and mobility research. First, the strong demographic-spatial relationships validate co-evolutionary theories of urban development, demonstrating quantifiable bidirectional influences between built environment and population characteristics. Second, the high prediction accuracy across demographic groups reveals the persistent spatial imprint of historical processes, supporting path-dependency theories in urban development.

The travel behavior results contribute to theories about residential self-selection versus built environment effects by demonstrating that spatial configurations encode sufficient information to predict mobility patterns, suggesting stronger built environment influences than purely preference-based models would predict \cite{mokhtarian2008examining}. However, the variable performance across transportation modes highlights the continuing importance of individual agency and choice within spatial constraints.

Our approach's superior demographic consistency compared to purely visual models addresses calls for more demographically-aware planning tools \cite{lucas2012transport}. By explicitly modeling demographic-spatial relationships, the framework supports more equitable planning processes that recognize how development patterns both reflect and reinforce social inequalities.

\subsection{Limitations and Future Directions}

While demonstrating strong performance, our model exhibits limitations reflecting broader challenges in spatial analysis. The reduced accuracy for certain demographic groups (particularly minority populations) may reflect the complex, non-spatial factors influencing residential patterns. Future research could incorporate additional contextual variables such as policy frameworks, economic conditions, or historical development patterns.

\subsection{Data Availability Statement}

The dataset is publicly available on \href{https://ieee-dataport.org/documents/
county-level-us-multi-modal-spatiotemporal-urban-growth-travel-behavior-dataset-2012-2023}
{IEEE DataPort} (DOI: \href{https://doi.org/10.21227/fn8z-8z82}{10.21227/fn8z-8z82}).  
The raw satellite imagery data are publicly accessible via Google Earth Engine.  
The raw demographic and travel behavior data are publicly available from the U.S. Census Bureau.

\section{Conclusion}
This study demonstrates that integrating demographic context into spatial prediction models produces substantial improvements in both visual accuracy and sociodemographic consistency. The proposed Demographics-Informed Neural Network (DINN) achieves superior performance with structural similarity (SSIM) of 0.8342 and demographic consistency loss of 0.14, representing significant advances over existing state-of-the-art approaches. These quantitative results provide strong evidence that combining satellite imagery with demographic data creates more accurate and realistic urban predictions than methods relying solely on visual information.
The findings validate established theories that emphasize the co-evolutionary relationship between built environments and social structures. Specifically, the results confirm that physical landscapes and demographic patterns develop together in predictable ways, with each influencing the other over time. This validation has important implications because it demonstrates that urban development follows systematic patterns that can be captured and modeled computationally. Furthermore, the study provides practical tools for more socially-aware planning by enabling planners to understand how proposed developments might affect different demographic groups and communities.
The bidirectional relationship between spatial patterns and demographic characteristics, quantified through the integrated architecture, offers new insights for addressing persistent urban challenges. These challenges include spatial mismatch, where job locations are disconnected from where people live; residential segregation, where different demographic groups remain separated across neighborhoods; and inequitable development, where infrastructure investments do not benefit all communities equally. By demonstrating that demographic patterns can be predicted from satellite images and vice versa, this research provides planners with tools to anticipate and potentially mitigate these problems before they become entrenched in the urban landscape.

\bibliographystyle{ieeetr}

\bibliography{MyBib}

\begin{thebibliography}{10}

\bibitem{gramlich94}
E.~M. Gramlich, ``Infrastructure investment: A review essay,'' {\em Journal of Economic Literature}, vol.~32, no.~3, pp.~1176--1196, 1994.

\bibitem{POKHAREL2023100817}
R.~Pokharel, L.~Bertolini, and M.~{te Brömmelstroet}, ``How does transportation facilitate regional economic development? a heuristic mapping of the literature,'' {\em Transportation Research Interdisciplinary Perspectives}, vol.~19, p.~100817, 2023.

\bibitem{SERDAR2022103452}
M.~Z. Serdar, M.~Koç, and S.~G. Al-Ghamdi, ``Urban transportation networks resilience: Indicators, disturbances, and assessment methods,'' {\em Sustainable Cities and Society}, vol.~76, p.~103452, 2022.

\bibitem{DANYO2025}
A.~Danyo, A.~Dontoh, and A.~Aboah, ``An improved resnet50 model for predicting pavement condition index (pci) directly from pavement images,'' {\em Road Materials and Pavement Design}, pp.~1--18, 2025.

\bibitem{su15108410}
M.~Chanieabate, H.~He, C.~Guo, B.~Abrahamgeremew, and Y.~Huang, ``Examining the relationship between transportation infrastructure, urbanization level and rural-urban income gap in china,'' {\em Sustainability}, vol.~15, no.~10, 2023.

\bibitem{WAN2024384}
J.~Wan, Q.~Xie, and X.~Fan, ``The impact of transportation and information infrastructure on urban productivity: Evidence from 256 cities in china,'' {\em Structural Change and Economic Dynamics}, vol.~68, pp.~384--392, 2024.

\bibitem{WANG201725}
D.~Wang and X.~Cao, ``Impacts of the built environment on activity-travel behavior: Are there differences between public and private housing residents in hong kong?,'' {\em Transportation Research Part A: Policy and Practice}, vol.~103, pp.~25--35, 2017.

\bibitem{wang2024deep}
Q.~Wang, S.~Wang, Y.~Zheng, H.~Lin, X.~Zhang, J.~Zhao, and J.~Walker, ``Deep hybrid model with satellite imagery: How to combine demand modeling and computer vision for travel behavior analysis?,'' {\em Transportation Research Part B: Methodological}, vol.~179, p.~102869, 2024.

\bibitem{denteh2025integrating}
E.~Denteh, A.~Danyo, J.~K. Asamoah, B.~A. Kyem, T.~Addai, and A.~Aboah, ``Integrating travel behavior forecasting and generative modeling for predicting future urban mobility and spatial transformations,'' {\em arXiv preprint arXiv:2503.21158}, 2025.

\bibitem{LIU2024}
L.~Liu, Y.~Li, D.~Gruyer, and M.~Tu, ``Non-linear relationship between built environment and active travel: A hybrid model considering spatial heterogeneity,'' {\em International Journal of Transportation Science and Technology}, 2024.

\bibitem{racheampong}
R.~Acheampong and E.~Silva, ``Land use-transport interaction modelling: A review of the literature and future research directions,'' {\em Journal of Land use and Transport}, vol.~8, pp.~1--28, 06 2015.

\bibitem{kim2024travel}
K.~Kim, D.~Byrd, and S.~Handy, ``Travel demand modeling and the assessment of environmental impacts: A literature review,'' {\em UC Davis: National Center for Sustainable Transportation}, 2024.

\bibitem{boyles2006comparison}
S.~Boyles, S.~V. Ukkusuri, S.~T. Waller, and K.~M. Kockelman, ``A comparison of static and dynamic traffic assignment under tolls: a study of the dallas-fort worth network,'' in {\em 85th Annual meeting of the transportation research board}, 2006.

\bibitem{vanetten2021multitemporalurbandevelopmentspacenet}
A.~V. Etten, D.~Hogan, J.~Martinez-Manso, J.~Shermeyer, N.~Weir, and R.~Lewis, ``The multi-temporal urban development spacenet dataset,'' 2021.

\bibitem{pugh2024rochester}
C.~Pugh, ``Rochester's inner loop freeway-to-boulevard project: A case study,'' 2024.

\bibitem{Sciara03072017}
G.-C.~S. and, ``Metropolitan transportation planning: Lessons from the past, institutions for the future,'' {\em Journal of the American Planning Association}, vol.~83, no.~3, pp.~262--276, 2017.

\bibitem{HORCHER2021100196}
D.~Hörcher and A.~Tirachini, ``A review of public transport economics,'' {\em Economics of Transportation}, vol.~25, p.~100196, 2021.

\bibitem{Mladenovic}
M.~Mladenovic and A.~Trifunovic, ``The shortcomings of the conventional four step travel demand forecasting process,'' {\em Journal of Road and Traffic Engineering}, 01 2014.

\bibitem{BIBRI2020100021}
S.~E. Bibri, J.~Krogstie, and M.~Kärrholm, ``Compact city planning and development: Emerging practices and strategies for achieving the goals of sustainability,'' {\em Developments in the Built Environment}, vol.~4, p.~100021, 2020.

\bibitem{BALSAS20151}
C.~J. Balsas, ``Sustainable transportation planning, a new academic specialization in the usa,'' {\em International Journal of Transportation Science and Technology}, vol.~4, no.~1, pp.~1--15, 2015.

\bibitem{abirami2024systematic}
S.~Abirami, M.~Pethuraj, M.~Uthayakumar, and P.~Chitra, ``A systematic survey on big data and artificial intelligence algorithms for intelligent transportation system,'' {\em Case Studies On Transport Policy}, p.~101247, 2024.

\bibitem{kyem2024advancing}
B.~A. Kyem, E.~K.~O. Denteh, J.~K. Asamoah, K.~A. Tutu, and A.~Aboah, ``Advancing pavement distress detection in developing countries: A novel deep learning approach with locally-collected datasets,'' {\em arXiv preprint arXiv:2408.05649}, 2024.

\bibitem{kyem2024pavecap}
B.~A. Kyem, E.~K.~O. Denteh, J.~K. Asamoah, and A.~Aboah, ``Pavecap: The first multimodal framework for comprehensive pavement condition assessment with dense captioning and pci estimation,'' {\em arXiv preprint arXiv:2408.04110}, 2024.

\bibitem{kyem2024weather}
B.~A. Kyem, J.~K. Asamoah, Y.~Huang, and A.~Aboah, ``Weather-adaptive synthetic data generation for enhanced power line inspection using stargan,'' {\em IEEE Access}, vol.~12, pp.~193882--193901, 2024.

\bibitem{kyem2025context}
B.~A. Kyem, J.~K. Asamoah, and A.~Aboah, ``Context-cracknet: A context-aware framework for precise segmentation of tiny cracks in pavement images,'' {\em arXiv preprint arXiv:2501.14413}, 2025.

\bibitem{asamoah2025saam}
J.~K. Asamoah, B.~A. Kyem, N.-D. Obeng-Amoako, and A.~Aboah, ``Saam-reflectnet: Sign-aware attention-based multitasking framework for integrated traffic sign detection and retroreflectivity estimation,'' {\em Expert Systems with Applications}, p.~128003, 2025.

\bibitem{dontoh2025visual}
A.~Dontoh, S.~Ivey, L.~Sirbaugh, A.~Danyo, and A.~Aboah, ``Visual dominance and emerging multimodal approaches in distracted driving detection: A review of machine learning techniques,'' {\em arXiv preprint arXiv:2505.01973}, 2025.

\bibitem{li2019}
X.~Li, Y.~Zhou, J.~Eom, S.~Yu, and G.~R. Asrar, ``Projecting global urban area growth through 2100 based on historical time series data and future shared socioeconomic pathways,'' {\em Earth's Future}, vol.~7, no.~4, pp.~351--362, 2019.

\bibitem{zhang2024characteristics}
X.~Zhang and H.~Han, ``Characteristics and factors influencing the expansion of urban construction land in china,'' {\em Scientific Reports}, vol.~14, no.~1, p.~16040, 2024.

\bibitem{karas2015highway}
D.~Karas, ``Highway to inequity: the disparate impact of the interstate highway system on poor and minority communities in american cities,'' {\em New Visions for Public Affairs}, vol.~7, no.~April, pp.~9--21, 2015.

\bibitem{LU19991}
X.~Lu and E.~I. Pas, ``Socio-demographics, activity participation and travel behavior,'' {\em Transportation Research Part A: Policy and Practice}, vol.~33, no.~1, pp.~1--18, 1999.

\bibitem{MWALE2022100683}
M.~Mwale, R.~Luke, and N.~Pisa, ``Factors that affect travel behaviour in developing cities: A methodological review,'' {\em Transportation Research Interdisciplinary Perspectives}, vol.~16, p.~100683, 2022.

\bibitem{GORELICK201718}
N.~Gorelick, M.~Hancher, M.~Dixon, S.~Ilyushchenko, D.~Thau, and R.~Moore, ``Google earth engine: Planetary-scale geospatial analysis for everyone,'' {\em Remote Sensing of Environment}, vol.~202, pp.~18--27, 2017.
\newblock Big Remotely Sensed Data: tools, applications and experiences.

\bibitem{cervero1997travel}
R.~Cervero and K.~Kockelman, ``Travel demand and the 3ds: Density, diversity, and design,'' {\em Transportation research part D: Transport and environment}, vol.~2, no.~3, pp.~199--219, 1997.

\bibitem{mieszkowski1993causes}
P.~Mieszkowski and E.~S. Mills, ``The causes of metropolitan suburbanization,'' {\em Journal of Economic perspectives}, vol.~7, no.~3, pp.~135--147, 1993.

\bibitem{florida2011creative}
R.~Florida, ``“the creative class”: from the rise of the creative class: And how it’s transforming work, leisure, community and everyday life (2002),'' in {\em The City Reader}, pp.~175--181, Routledge, 2011.

\bibitem{massey2019american}
D.~S. Massey and N.~A. Denton, ``American apartheid: Segregation and the making of the underclass,'' in {\em Social Stratification, Class, Race, and Gender in Sociological Perspective, Second Edition}, pp.~660--670, Routledge, 2019.

\bibitem{arthur1994increasing}
W.~B. Arthur, {\em Increasing returns and path dependence in the economy}.
\newblock University of michigan Press, 1994.

\bibitem{ewing2001travel}
R.~Ewing and R.~Cervero, ``Travel and the built environment: a synthesis,'' {\em Transportation research record}, vol.~1780, no.~1, pp.~87--114, 2001.

\bibitem{hagerstrand1970people}
T.~H{\"a}gerstrand, ``What about people in regional science,'' {\em Transport Sociology: Social aspects of transport planning}, pp.~143--158, 1970.

\bibitem{newman1989gasoline}
P.~W. Newman and J.~R. Kenworthy, ``Gasoline consumption and cities: a comparison of us cities with a global survey,'' {\em Journal of the American planning association}, vol.~55, no.~1, pp.~24--37, 1989.

\bibitem{saelens2003environmental}
B.~E. Saelens, J.~F. Sallis, and L.~D. Frank, ``Environmental correlates of walking and cycling: findings from the transportation, urban design, and planning literatures,'' {\em Annals of behavioral medicine}, vol.~25, no.~2, pp.~80--91, 2003.

\bibitem{mokhtarian2008examining}
P.~L. Mokhtarian and X.~Cao, ``Examining the impacts of residential self-selection on travel behavior: A focus on methodologies,'' {\em Transportation Research Part B: Methodological}, vol.~42, no.~3, pp.~204--228, 2008.

\bibitem{lucas2012transport}
K.~Lucas, ``Transport and social exclusion: Where are we now?,'' {\em Transport policy}, vol.~20, pp.~105--113, 2012.

\end{thebibliography}

\section{Biography Section}

\begin{IEEEbiography}[{\includegraphics[width=1in,height=1.25in,clip,keepaspectratio]{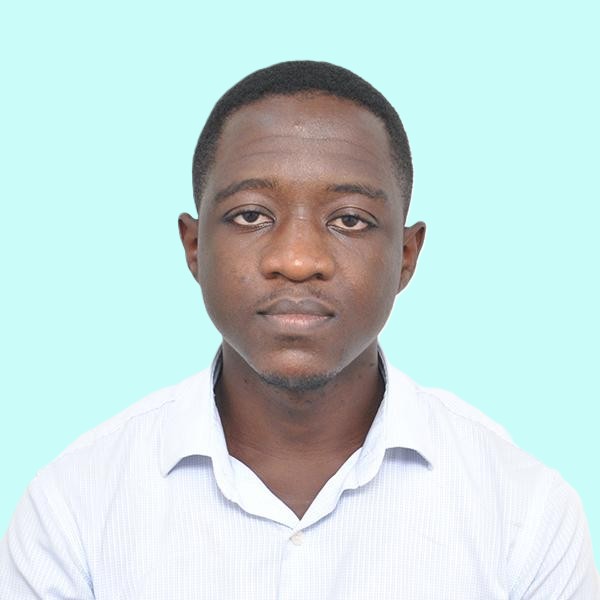}}]{Eugene Kofi Okrah Denteh} Eugene Kofi Okrah Denteh is a first year Ph.D. student and Research Assistant at North Dakota State University (NDSU), where he contributes to the SMART Lab, an interdisciplinary research group focused on sustainable mobility and advanced transportation systems. His academic journey began with a Bachelor of Science in Civil Engineering from Kwame Nkrumah University of Science and Technology (KNUST) in Ghana.
Denteh's research interests encompass transportation planning, travel demand modeling, congestion management, pavement and asset management, and transportation safety. He is particularly interested in integrating artificial intelligence (AI) and big data analytics into transportation engineering to develop innovative solutions for urban mobility challenges. One of his notable contributions includes the development of a two-stage deep learning framework that combines a Temporal Fusion Transformer for travel behavior forecasting with a Generative Adversarial Network (GAN) for spatial prediction. Through his work, Eugene Denteh aims to bridge the gap between traditional civil engineering practices and modern AI-driven approaches, contributing to the evolution of smart cities and efficient transportation systems.
\end{IEEEbiography}

\begin{IEEEbiography}[{\includegraphics[width=1in,height=1.25in,clip,keepaspectratio]{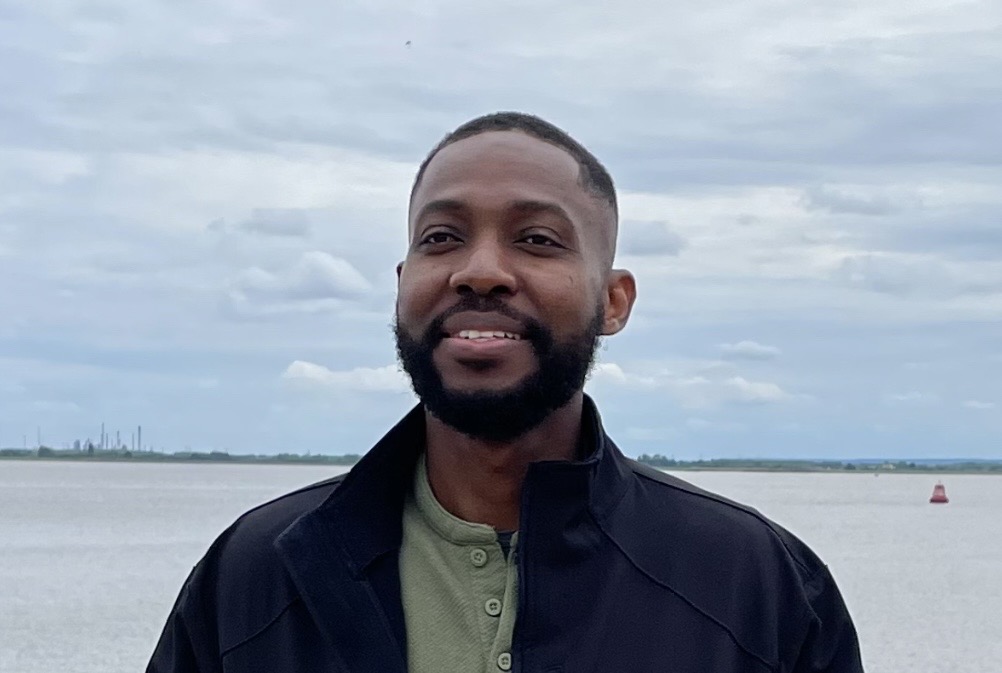}}]{Andrews Danyo} Andrews Danyo is a Civil Engineer passionate about innovation and smart initiatives. He also has a background in construction management, artificial intelligence, and data science. His work focuses on creating smarter transportation systems and urban infrastructure that serves people, not just statistics. What drives Andrews is seeing how intelligent systems can solve real-world infrastructure challenges or develop sustainable solutions that adapt to human needs. He brings practical knowledge of how things get built and the vision for how technology can make them work better for everyone.
\end{IEEEbiography}
\begin{IEEEbiography}[{\includegraphics[width=1in,height=1.25in,clip,keepaspectratio]{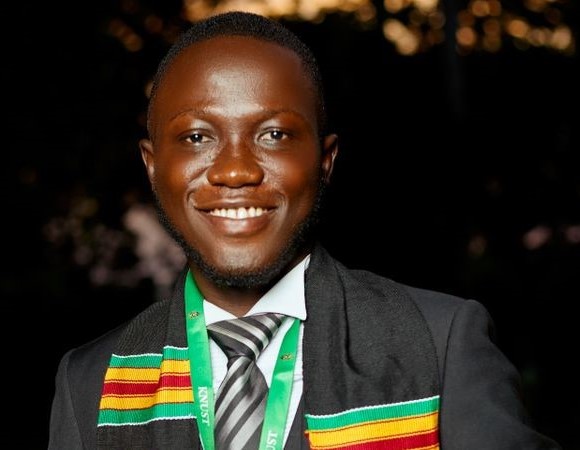}}]{Joshua Kofi Asamoah} Joshua earned his Bachelor's degree in Civil Engineering from Kwame Nkrumah University of Science and Technology in Ghana in 2023. Currently, Joshua is a first-year doctoral student in the Civil Engineering program with a concentration in Transportation Engineering at North Dakota State University, where he also serves as a Graduate and Teaching Assistant. Under the mentorship of Professor Armstrong Aboah at the SMART lab, Joshua's research interests lie in the realms of machine learning, deep learning, computer vision, and Internet of Things (IoT), with a particular emphasis on their applications in autonomous navigation and perception. His work aims to leverage these cutting-edge technologies to enhance the capabilities of autonomous systems and improve their ability to navigate and perceive their surroundings effectively.
At present, Joshua is engaged in a research project that focuses on predicting lane intentions and vehicle trajectories using Naturalistic driving data. This project employs advanced computer vision and machine learning techniques to analyze real-world driving scenarios and develop predictive models that can anticipate the behavior of surrounding vehicles. Additionally, he is exploring the integration of IoT technologies to augment the perception capabilities of autonomous systems, enabling more efficient and safer navigation in complex environments.
\end{IEEEbiography}
\begin{IEEEbiography}[{\includegraphics[width=1in,height=1.25in,clip,keepaspectratio]{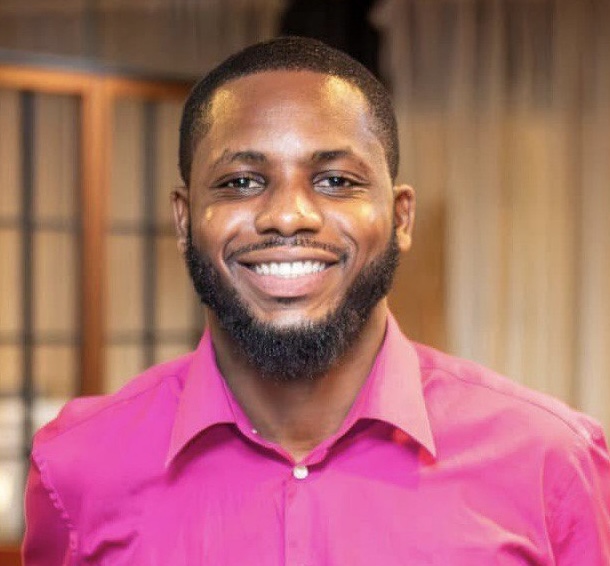}}]{Blessing Agyei Kyem} obtained his B.S. in Civil Engineering from the Kwame Nkrumah University of Science and Technology, Ghana, in 2023. He worked as a Data Scientist at Bismuth Technologies from 2022-2023. Currently, Blessing is a first-year Ph.D. student in Civil Engineering with a concentration in Transportation Engineering and serves as a Graduate and Teaching Assistant at North Dakota State University. Under the guidance of Professor Armstrong Aboah at the SMART lab, Blessing's research interests revolve around machine learning, deep learning, computer vision, natural language processing, and their applications in Naturalistic driving, Pavement Asset Management, Intelligent Transportation Systems, and Connected and Autonomous Vehicles. Additionally, he is exploring the application of multi-modal models for Pavement Asset Management and Transportation leveraging multiple data sources for improved decision-making. One of his novel research works is PaveCap, a novel multimodal framework that uses YOLOv8 and SAM for object detection and segmentation, ultimately predicting PCI and generating textual descriptions of pavement conditions to aid in infrastructure management. 
\end{IEEEbiography}
\begin{IEEEbiography}
[{\includegraphics[width=1in,height=1.25in,clip,keepaspectratio]{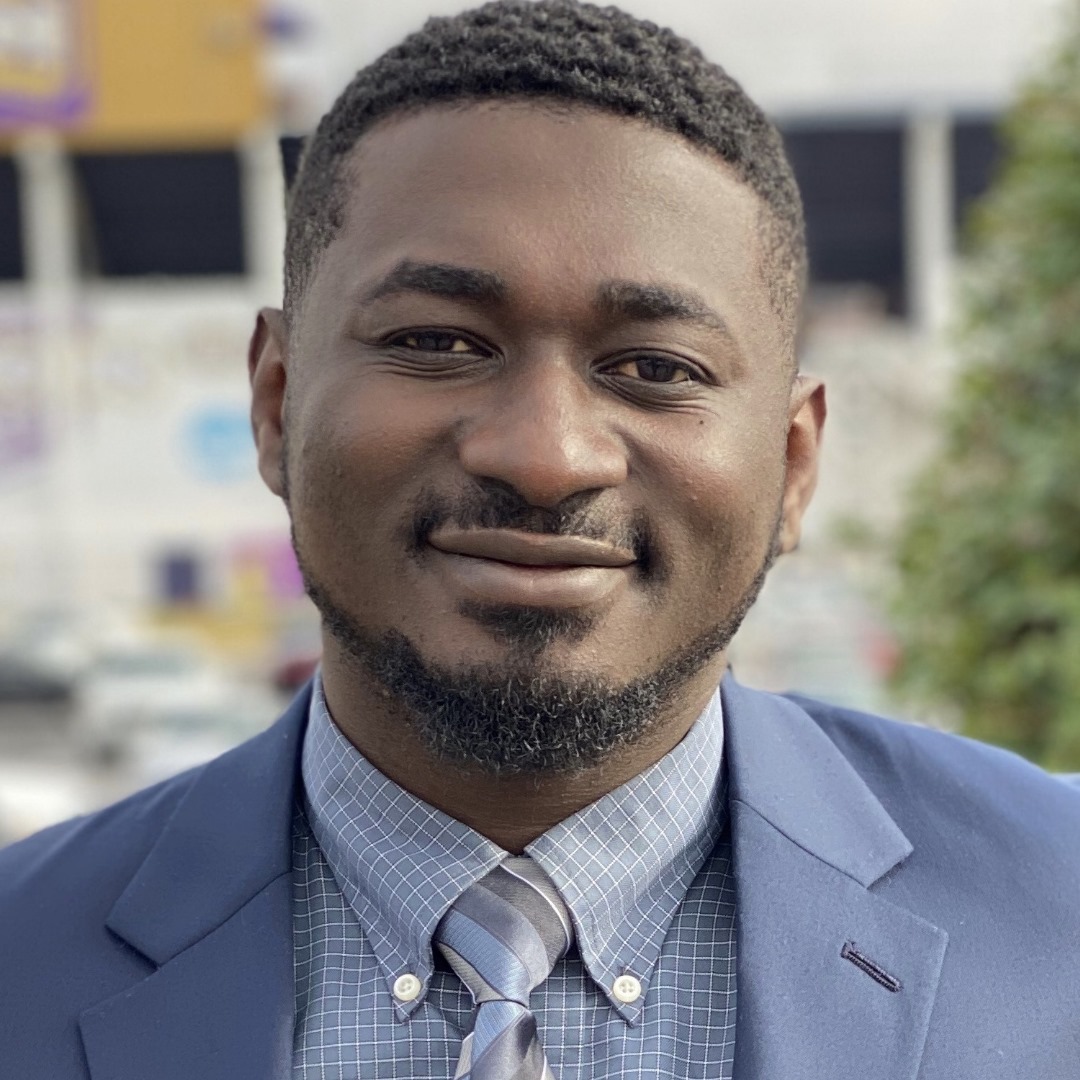}}]{Armstrong Aboah} (M'24) was born in Kumasi, Ghana. He received the B.S. degree in civil engineering from the Kwame Nkrumah University of Science and Technology,
in 2017, the M.S. degree in transportation engineering from Tennessee Technological University, in 2019, and the Ph.D. degree in transportation
engineering, with a focus in computer vision and machine learning from the University of Missouri, Columbia, MO, USA, in 2022. His papers have been published at top venues, including CVPR, NeuIPS, and the Journal of Transportation Engineering. His papers have received more than 430 citations. He has secured competitive research funding from various agencies, including projects on extracting insights from naturalistic driving data and factors influencing transportation network company usage. His research interests include computer vision, transportation sensing, big data analytics, and deep learning, with a focus on intelligent transportation systems. He has authored over 15 peer-reviewed publications in these areas. He is a leading researcher on vision-based traffic anomaly detection and has published one of the first papers applying deep learning and decision trees to this problem. His paper on real-time multi-class helmet violation detection using few-shot learning techniques has also been impactful. Other novel research contributions include smartphone-based pavement roughness estimation using deep learning, prediction of bus delays across multiple
routes, and automated retail checkout using computer vision.
\end{IEEEbiography}

\end{document}